\documentclass[11pt]{article}
\usepackage{geometry}
\usepackage{coling2020}
\usepackage{times}
\usepackage{url}
\usepackage{latexsym}
\usepackage{microtype}
\usepackage{times}
\usepackage[table,xcdraw]{xcolor}
\usepackage{latexsym}
\usepackage{url}
\usepackage{amsmath,amssymb,amsfonts}
\usepackage{algorithmic}
\usepackage{graphicx}
\usepackage{textcomp}
\usepackage{float}
\usepackage{enumitem}
\usepackage{amsmath,amssymb,amsfonts}
\usepackage[ruled,vlined]{algorithm2e}
\usepackage{algorithmic}
\usepackage{graphicx}
\usepackage{textcomp}
\usepackage{multirow}
\usepackage{csquotes}
\usepackage{subfig}

\colingfinalcopy 

\title{Investigating the Effect of Intraclass Variability in Temporal Ensembling}

\author{Siddharth Vohra\textsuperscript{1}\thanks{\ \ Work performed while interning at Hitachi R\&D India} , Manikandan Ravikiran\textsuperscript{2} \\
  \textsuperscript{1}University of California, San Diego\\
  \textsuperscript{2}Research \& Development Center, Hitachi India Pvt Ltd., Bangalore, India\\
  {\tt sivohra@ucsd.edu} \\
  {\tt manikandan@hitachi.co.in}}
  
\date{}

\begin{document}
\maketitle
\begin{abstract}
Temporal Ensembling is a semi-supervised approach that allows training deep neural network models with a small number of labeled images. In this paper, we present our preliminary study on the effect of intraclass variability on temporal ensembling, with a focus on seed size and seed type, respectively. Through our experiments we find that (a) there is a significant drop in accuracy with datasets that offer high intraclass variability, (b) more seed images offer consistently higher accuracy across the datasets, and (c) seed type indeed has an impact on the overall efficiency, where it produces a spectrum of accuracy both lower and higher. Additionally, based on our experiments, we also find KMNIST to be a competitive baseline for temporal ensembling.   
\end{abstract}

\section{Introduction}\label{intro}

Deep neural networks have seen broad applications across vision speech and language in recent times. Yet this success is contingent on acquiring a large number of labeled datasets, which is expensive and time-consuming. Further, labeling is mostly manual, done by humans, due to its higher meticulousness. Recently, to address this concern of manual labeling variety of approaches have been designed, including Semi-Supervised learning algorithms \cite{Gordon2017BayesianSL,Laine2017TemporalEF,Lee2013PseudoLabelT}, which typically proffer with higher results with a small number of labeled examples (seeds). Notable among this is the Temporal Ensembling \cite{Laine2017TemporalEF}, which uses an ensemble of the earlier outputs of a neural network as an unsupervised target label and achieved high accuracy on SVHN and CIFAR-10 with just {500} and {4000} labeled samples with both naturally offering lower intraclass variances. Besides, to the best of our knowledge, there is no explicit study of temporal ensembling in the context of datasets with large intraclass variability. As such, in this work, we attempt to investigate this gap by answering the following research questions.

\begin{itemize}
\itemsep0em
    \item[--] \textbf{RQ1:} Does intraclass variability impact the accuracy of temporal ensembling? Here the intention is to check (a) how accuracy varies and (b) if there is any unique observable behavior with temporal ensembling under different intraclass variances. The assumption here being that the intraclass variability is a spectrum with a range of low to high intraclass variation. To this end, we experiment on datasets of  Fashion-MNIST \cite{Xiao2017FashionMNISTAN} and KMNIST \cite{Clanuwat2018DeepLF} to find that there is a sheer drop in performance when using temporal ensembling.
    \item[--] \textbf{RQ2:} Under settings of intraclass variability, how does seed size impact temporal ensembling?. Here we hypothesize and verify that upon increasing seed size, there is an improvement in performance.
    \item[--] \textbf{RQ3:} What's the effect of seed selection on temporal ensembling? More specifically, we see if the diversity of seeds have an impact on results. The preliminary experimental results show that performance is lower with some category of seeds over others. 
\end{itemize}

The rest of the paper is organized as follows. In section \ref{related_work}, we review related research in semi-supervised learning. In section \ref{temporal}, we introduce the temporal ensembling approach in brief. In section \ref{data_exp}, we present dataset, experimental setup, and answer each of the research questions with analysis in section \ref{exps}.  Finally, we conclude in section \ref{futurework} with possible implication on future works.

\section{Related Work}\label{related_work}
Semi-Supervised learning has seen an extensive assortment of works originating with pseudo-labeling \cite{Lee2013PseudoLabelT}, which assigns pseudo-labels to unlabelled data using the model forecasts at each epoch and training on both labeled and pseudo-labeled data. Then there are variational autoencoders \cite{Kingma2014AutoEncodingVB}, which uses a generative deep learning model \cite{Kingma2014SemisupervisedLW} for semi-supervised learning. Similarly, there are Ladder Networks \cite{Rasmus2015SemisupervisedLW}, which uses noise to the input of each layer in the neural network and combines it with denoising function. Ladder network combines denoising loss with the supervised loss during final training. More recently, there is Virtual Adversarial Training (VAT) \cite{Miyato2015DistributionalSW}, which uses perturbations generated by adversarial learning, Temporal Ensembling \cite{Laine2017TemporalEF}, which is like pseudo-labeling, except it matches the output of previous models and Mean-Teacher \cite{Tarvainen2017MeanTA}, which addresses the drawback of the storage requirement of temporal ensembling by using an exponential moving average.  There are many other notable works, including works on co-training and feature learning \cite{Blum1998CombiningLA,Sindhwani2005ACA}, and graph-based methods \cite{Ng2018BayesianSL}, metric-learning based methods \cite{Yu2017DeepMD} and sampling noisy labels \cite{Vahdat2017TowardRA,Ravikiran2020HitachiAS}.  In this work, we concentrate on temporal ensembling due to its simplicity and study it in the context of datasets with diverse intraclass variability.

\section{Temporal Ensembling}\label{temporal}
Temporal Ensembling is an enhancement of the $\Pi$-model \cite{Laine2017TemporalEF} and is based on the idea of self-ensembling where it ensembles earlier outputs as an unsupervised target (See Equations \ref{eq1}-\ref{eq3}). This process of self-ensembling is analogous to that of pseudo-labels. More specifically, in temporal ensembling training is done with dataset under different augmentations and dropout regularization, which presents the network to acquire noise-invariant features. 

\begin{equation}\label{eq1}
    l_{B}(z, \tilde{z}, y) = masked\_crossentropy(z, y) + w(t) * MSE(z, \tilde{z})
\end{equation}

\begin{equation}\label{eq2}
    masked\_crossentropy(z, y) = - \frac{1}{\mid B \cap L \mid} \sum_{i \in (B \cap L)}{\log{z_{i}[y_{i}]}}
\end{equation}

\begin{equation}\label{eq3}
    MSE(z, \tilde{z}) = \frac{1}{C \mid B \mid} \sum_{i \in B}{\mid \mid z_{i} - \tilde{z}_{i} \mid \mid ^{2}}
\end{equation}
\noindent
This, in turn, allows the neural network to not shift its predictions in slightly modified variants of the same inputs. After every training epoch, the ensemble outputs refresh with the current new prediction and the previous ensemble prediction through the exponential moving average (EMA) method shown in Equation \ref{eq4}. Also, to generate the input training target  $\tilde{z}$ as shown in equation \ref{eq5}, temporal ensembling corrects the startup bias of $Z$. More details on temporal ensembling can be found in \cite{Laine2017TemporalEF} and \cite{ferret_2018}.

\begin{equation}\label{eq4}
    Z = \alpha  Z + (1 - \alpha) z 
\end{equation}

\begin{equation}\label{eq5}
    \tilde{z} = \frac{Z}{1 - \alpha^{t}}
\end{equation}

\section{Experimental Setup}
\label{data_exp}

To answer the research questions from section \ref{intro} in this work, we compare and contrast performance of temporal ensembling on various datasets under varying settings of labeled samples. This section provides an brief overview of the datasets (section \ref{datasets}) and parameter settings (section \ref{param}) used.

\subsection{Dataset}\label{datasets}
For this work, we use the datasets, which are presented in Table \ref{datatable}. More details of datasets are presented inplace across sections \ref{5.1}-\ref{5.3} when necessary.

\begin{table}[!htb]
\centering
\scalebox{1}{
\begin{tabular}{|c|c|c|c|c|}
\hline
\rowcolor[HTML]{CCCCCC} 
\textbf{Dataset}       & \textbf{Image Size}    & \textbf{Train Size}  & \textbf{Test Size}   & \textbf{Description}             \\ \hline
\textbf{MNIST}         & \multirow{3}{*}{28x28} & \multirow{3}{*}{60k} & \multirow{3}{*}{10k} & Handrwitten characters of 0-9    \\ \cline{1-1} \cline{5-5} 
\textbf{KMNIST}        &                        &                      &                      & Handwritten Kuzushiji Characters \\ \cline{1-1} \cline{5-5} 
\textbf{Fashion-MNIST} &                        &                      &                      & Clothing images from Zolando.com \\ \hline
\end{tabular}}
\caption{Summary of dataset and its characteristics used experiments \ref{5.1}-\ref{5.3}.}
\label{datatable}
\end{table}

\subsection{Parameter Settings}\label{param}
The various settings employed to answer each of the RQ's are as described. Besides, all the RQ's employ standard settings for most of the parameters, as shown in Table \ref{table22} below. 

\begin{itemize}
\itemsep0em
    \item \textbf{RQ1:} For analysis of RQ1, we train and test temporal ensembling models with all the three datasets, under common parameter settings from Table \ref{table22} with 300 and 500 epochs respectively.
    \item \textbf{RQ2:} For RQ2, we use parameters from Table \ref{table22} and vary labeled examples in range 100-500.
    \item \textbf{RQ3:} For RQ3, in addition to setting mentioned in Table \ref{table22}, we run experiments with 10 different randomly sampled seeds.
\end{itemize}

\begin{table}[!htb]
\centering
\scalebox{1}{
\begin{tabular}{|c|c|}
\hline
\rowcolor[HTML]{CCCCCC} 
\textbf{Hyperparameters}              & \textbf{Values} \\ \hline
\textbf{Dropout}                 & 0.5             \\ \hline
\textbf{Standard Deviation}      & 0.15            \\ \hline
\textbf{Feature maps in Conv\_1} & 16              \\ \hline
\textbf{Feature maps in Conv\_2} & 32              \\ \hline
\textbf{Weight Normalization}    & True            \\ \hline
\textbf{Learning Rate}           & 0.002           \\ \hline
\textbf{Beta}                    & 0.99            \\ \hline
\textbf{Batch Size}              & 100             \\ \hline
\textbf{Alpha}                   & 0.6             \\ \hline
\textbf{Data Normalization}              & channel wise    \\ \hline
\end{tabular}}
\caption{Hyperparameter settings used for experiments in sections \ref{5.1}-\ref{5.3}.}
\label{table22}
\end{table}

\section{Experiments}\label{exps}

\subsection{RQ1: Performance of Temporal Ensembling under Intraclass Variability}\label{5.1}

The first research question asks how \textit{intraclass variability influences the accuracy using temporal ensembling}. To answer this, the accuracy score is reported on the datasets of MNIST, KMNIST, and Fashion-MNIST, respectively. The reason behind selecting these datasets is that all of them are grayscale with the same image size and proportion of images, but KMNIST and Fashion-MNIST offer wide intraclass variability. None of these datasets require any size normalization, thus making results comparable across the datasets.

\begin{table}[!htb]
\centering
\scalebox{1}{
\begin{tabular}{|c|c|c|c|}
\hline
\rowcolor[HTML]{CCCCCC}
\textbf{Epochs}       & \textbf{100} & \textbf{300} & \textbf{500} \\ \hline

 \textbf{MNIST}        & 93.27$\pm$2.104  & 97.21$\pm$0.808 & 97.43$\pm$1.069 \\ \hline
\textbf{KMNIST}       & 50.05$\pm$20.07  & 49.90$\pm$20.20  & 60.66$\pm$3.700  \\ \hline
\textbf{Fashion-MNIST} & 74.14$\pm$1.580 & 73.95$\pm$0.975 & 71.31$\pm$2.443 \\ \hline
\end{tabular}}
\caption{Accuracy of Temporal Ensembling on MNIST, KMNIST and Fashion-MNIST. All the results are with seed size of 100 and five episodes of training.}
\label{table1}
\end{table}

\begin{figure}[!ht]
    \centering
    
    \subfloat[Supervised Loss]{{\includegraphics[width=10cm]{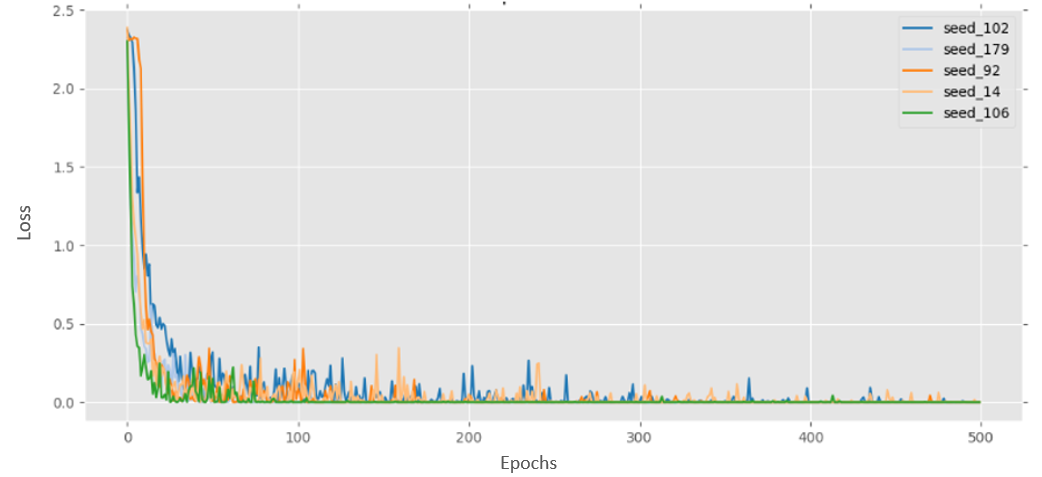} }}%
    \qquad
    \subfloat[Unsupervised Loss]{{\includegraphics[width=10cm]{MNIST_Supervised_loss.PNG} }}%
    \caption{Training loss behavior of MNIST using Temporal Ensembling}%
    \label{figmnist}%
\end{figure}

\begin{figure}[!ht]
    \centering
    \subfloat[Supervised Loss]{{\includegraphics[width=10cm]{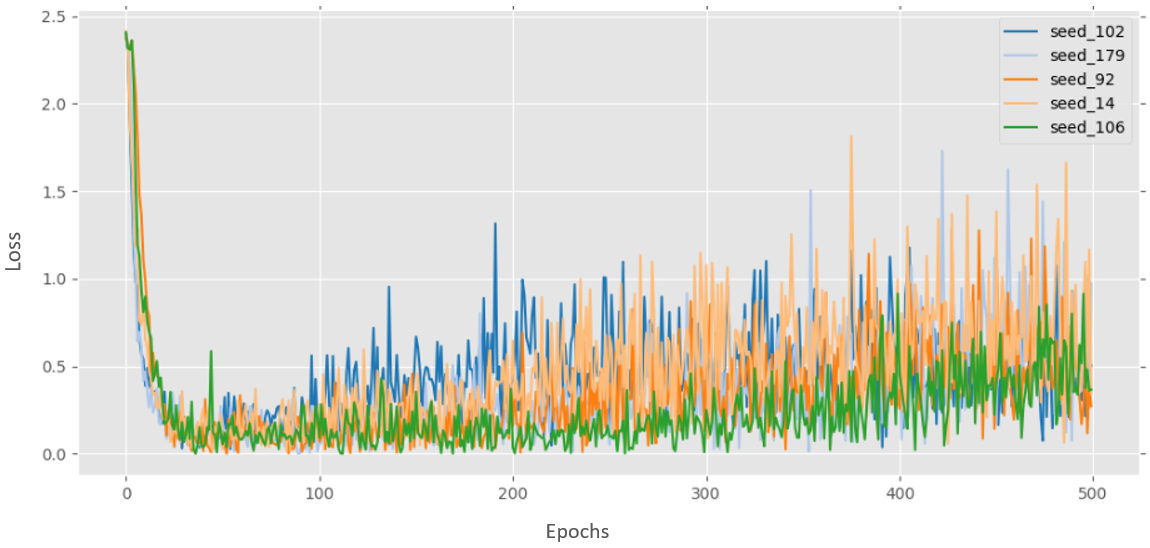} }}%
    \qquad
    \subfloat[Unsupervised Loss]{{\includegraphics[width=10cm]{KMNIST_Supervised_loss.PNG} }}%
    \caption{Training loss behavior of KMNIST using Temporal Ensembling}%
    \label{figkmnist}%
\end{figure}

\begin{figure}[!ht]
    \centering
    {
    \subfloat[Supervised Loss]{{\includegraphics[width=10cm]{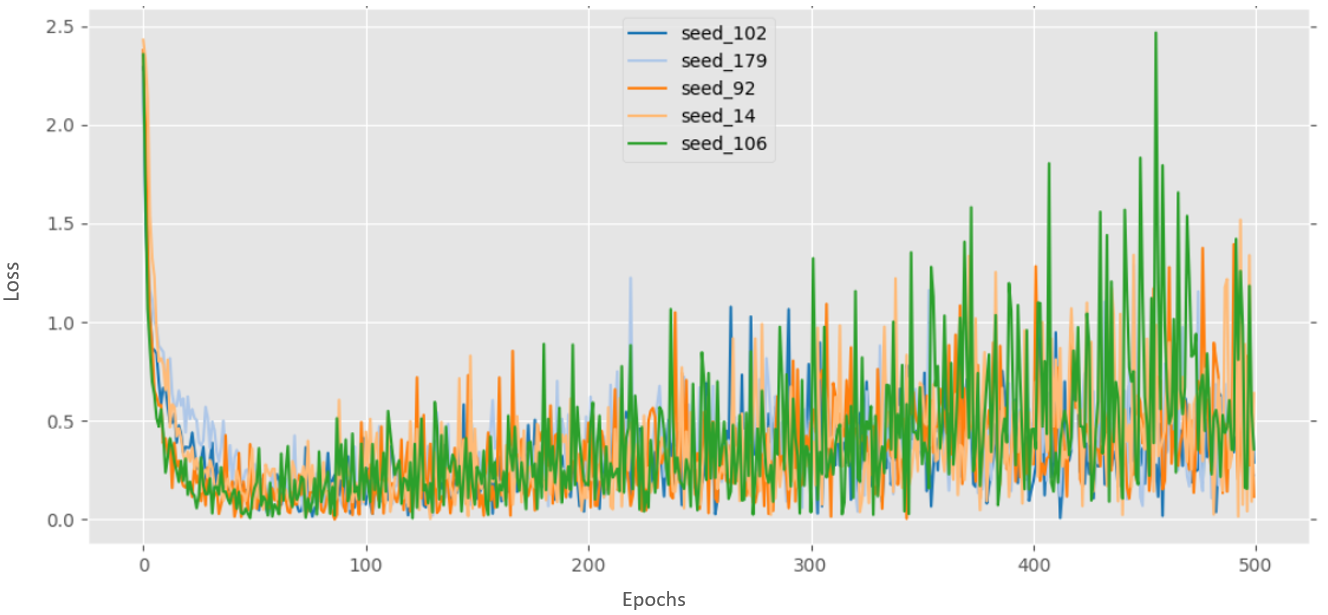} }}%
    \qquad
    \subfloat[Unsupervised Loss]{{\includegraphics[width=10cm]{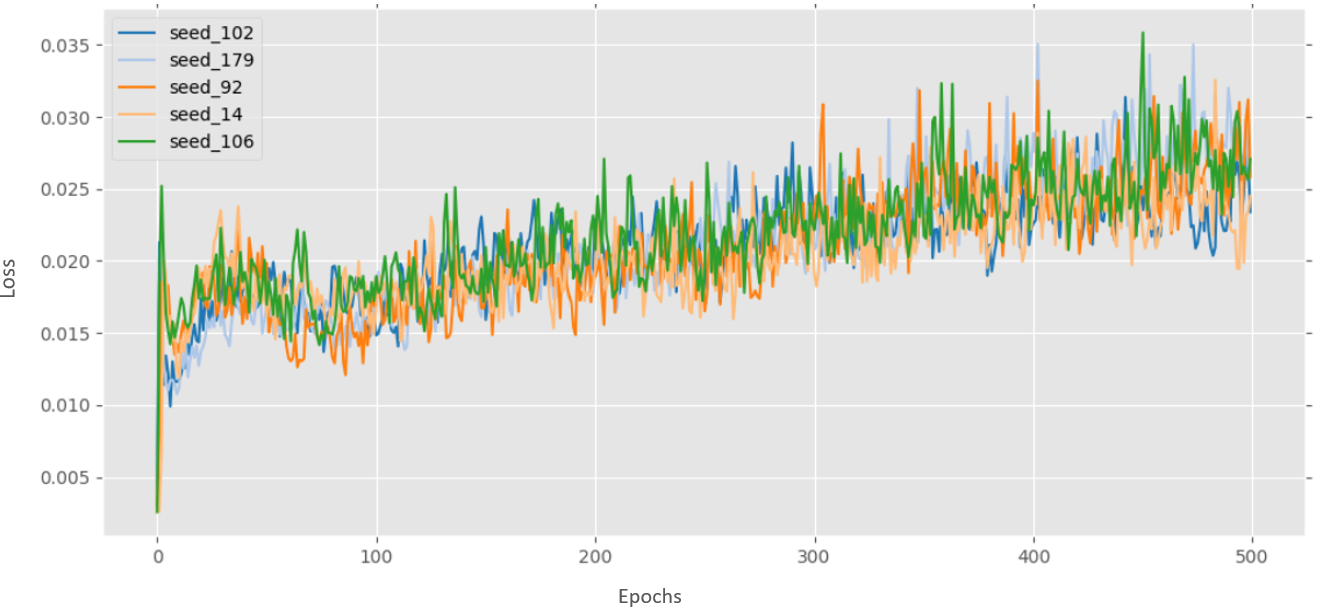} }}}%
    \caption{Training loss behavior of Fashion-MNIST using Temporal Ensembling}%
    \label{figFashion-MNIST}%
\end{figure}

Besides, as mentioned in section \ref{param}, here we trained temporal ensembling for 100, 300, and 500 epochs, respectively. The consolidated results so obtained are as shown in Table \ref{table1}. Also, we ran five episodes of training with varying seed sampling. Hence the results presented in Table \ref{table1} are averaged with a net standard deviation of accuracy. Detailed episode level results are in section \ref{rq1appendix}. Comparing the results of different datasets, we can see that MNIST gives the highest accuracy, followed by Fashion-MNIST and KMNIST, respectively. Besides, MNIST has faster convergence and stagnation in the loss (See Figure \ref{figmnist}), while KMNIST (Figure \ref{figkmnist}) and Fashion-MNIST (Figure \ref{figFashion-MNIST}) show delay in convergence time and increase in loss values with higher epochs for both supervised and unsupervised components. We believe such behavior is due to two reasons, namely (i) similarity (low variance) in train-test samples and samples within same classes for MNIST compared to KMNIST and Fashion-MNIST and (ii) unsupervised labels getting biased towards one of the classes, and no updates in the subsequent iterations of temporal ensembling.  However, more experiments are warranted to validate both (i) and (ii).

\noindent

Further, from Figure \ref{figmnist} and \ref{figkmnist}, we can see that such behavior is consistent across episodes with different seed samples for both KMNIST and Fashion-MNIST.  Besides, the results with KMNIST and Fashion-MNIST with some of the seeds differ by a large margin (See Tables \ref{tablerq1exps1.1}-\ref{tablerq1exps1.3} in Appendix).  Overall, analysis of temporal ensembling here shows that it is less tuned for datasets that have higher intraclass variances and might, therefore, have higher performances across varying datasets that do not match such criteria. Overall to summarize, our findings are as follows. 

\noindent
\begin{itemize}
\itemsep0em
\item Accuracy score is highest for MNIST (97.21\%), and among datasets that offer intraclass variance, there is a sheer drop in performance with KMNIST producing 60.66\% and Fashion-MNIST showing 71.31\%. We believe such a drop is due to intraclass variances and variance between train-test samples, both of which require further analysis.
\item Unlike MNIST, both KMNIST and Fashion-MNIST present with a unique behavior of lack of convergence for unsupervised loss and instead show an increase at higher epochs. This raises a question that the unsupervised labels may be biased towards a specific set of class(es) and aren't changing across iterations.  At the same time, it can also be conjectured that this could be due to the selection of learning rate or any other components of the temporal ensembling algorithm. Validating these requires a more detailed analysis of temporal ensembling and its components. 
\item Temporal ensembling is less suitable for datasets that offer higher intraclass variances, as evident from results. As such, it is not directly usable despite the similarity in size and proportion of datasets.
\end{itemize}

\subsection{RQ2: Effect of Seed Size on Temporal Ensembling under Intraclass Variability}\label{5.2}

Previously in section \ref{5.1}, we analyzed the impact of intraclass variance on temporal ensembling by studying performance on the MNIST,  KMNIST, and Fashion-MNIST datasets. However, for all the experiments in section \ref{5.1}, we maintained a uniform seed size of 100 labeled samples. Besides, in Figures \ref{figkmnist} and \ref{figFashion-MNIST}, we saw that seeds indeed have an impact on the overall results. As such, in this section, we shall investigate the effect of seed size, i.e., the number of labeled samples. More specifically, for RQ2, we vary seed size in the range of 100-500 and analyze the performance of models trained for both 300 and 500 epochs, respectively.

\begin{table}[!htb]
\centering
\scalebox{1}{
\begin{tabular}{|c|c|c|c|c|c|}
\hline
\rowcolor[HTML]{CCCCCC}
\cellcolor[HTML]{CCCCCC}\textbf{Seed   Size}  & \textbf{100} & \textbf{200} & \textbf{300} & \textbf{400} & \textbf{500} \\ \hline
 \textbf{MNIST}        & 97.21$\pm$0.8082 & 96.93$\pm$1.62   & 97.82$\pm$0.2128 & 97.41$\pm$.3285  & 97.37$\pm$0.2679 \\ \hline
 \textbf{KMNIST}       & 49.90$\pm$20.20  & 66.79$\pm$3.501  & 70.02$\pm$1.9912 & 72.24$\pm$2.3667 & 75.35$\pm$1.5323 \\ \hline
 \textbf{Fashion-MNIST} & 73.95$\pm$0.9755 & 76.71$\pm$1.524  & 78.7$\pm$1.1386  & 80.40$\pm$0.6780 & 81.53$\pm$1.036  \\ \hline
\end{tabular}}
\caption{Accuracy of Temporal Ensembling on MNIST, KMNIST and Fashion-MNIST with training for 300 epochs and 5 episodes.}
\label{table2_1}
\end{table}

\begin{table}[!htb]
\centering
\begin{tabular}{|c|c|c|c|c|c|}
\hline
\rowcolor[HTML]{CCCCCC}
\textbf{Seed   Size}  & \textbf{100} & \textbf{200} & \textbf{300} & \textbf{400} & \textbf{500} \\ \hline
\textbf{MNIST}        & 97.43$\pm$1.0698 & 97.53$\pm$0.7756 & 97.71$\pm$0.2838 & 97.93$\pm$0.1158 & 97.35$\pm$0.5745 \\ \hline
\textbf{KMNIST}       & 60.66$\pm$3.700  & 69.99$\pm$1.4461 & 70.11$\pm$3.0019 & 74.27$\pm$1.2281 & 75.36$\pm$1.6777 \\ \hline
\textbf{Fashion-MNIST} & 71.31$\pm$2.4430 & 76.81$\pm$0.8309 & 79.92$\pm$0.9292 & 80.68$\pm$0.9838 & 80.40$\pm$0.8048 \\ \hline
\end{tabular}
\caption{Accuracy of Temporal Ensembling on MNIST, KMNIST and Fashion-MNIST with training for 500 epochs and 5 episodes.}
\label{table2_2}
\end{table}

\begin{figure}[]
    \centering
    \subfloat[Supervised Loss]{{\includegraphics[width=10cm]{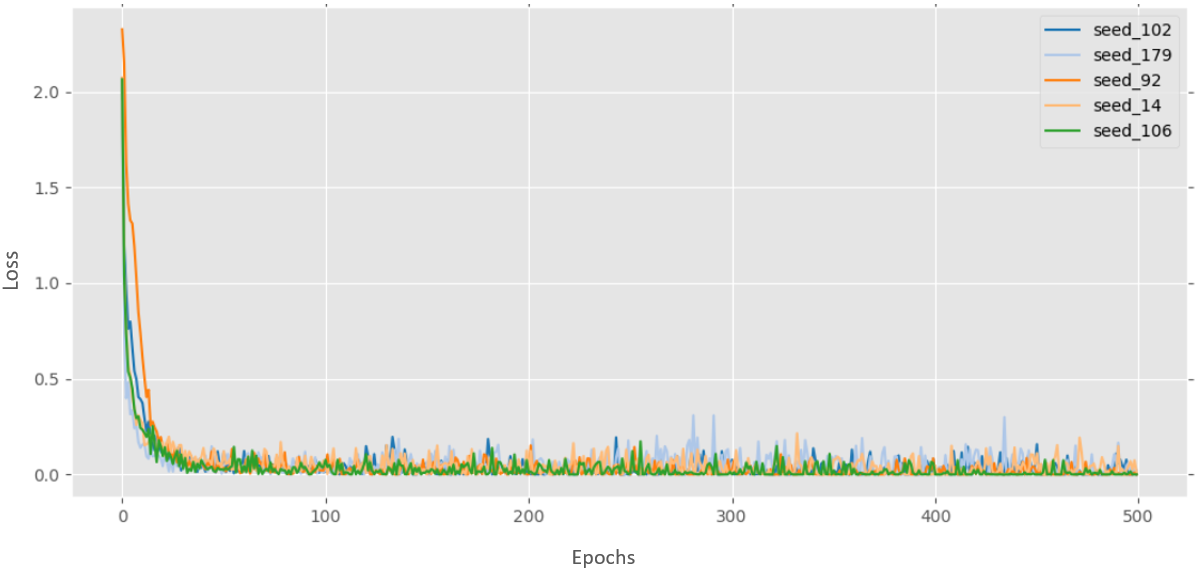} }}%
    \qquad
    \subfloat[Unsupervised Loss]{{\includegraphics[width=10cm]{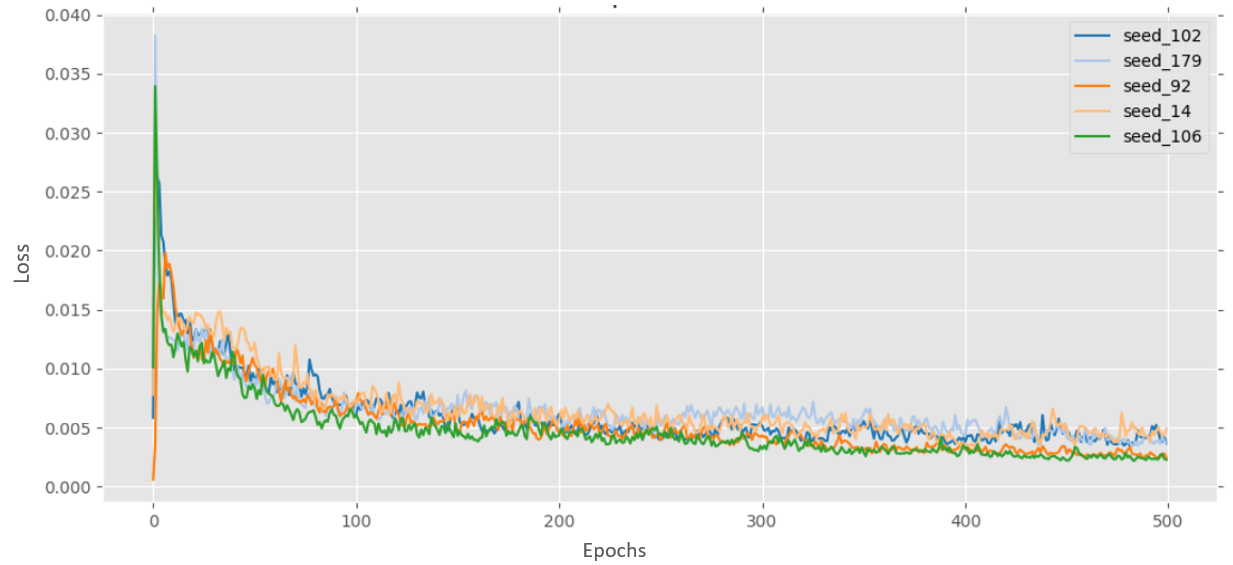} }}%
    \caption{Training loss behavior of MNIST using Temporal Ensembling with 300 Seeds}%
    \label{figmnist_seeds_300}%
\end{figure}

\begin{figure}[]
    \centering
    \subfloat[Supervised Loss]{{\includegraphics[width=10cm]{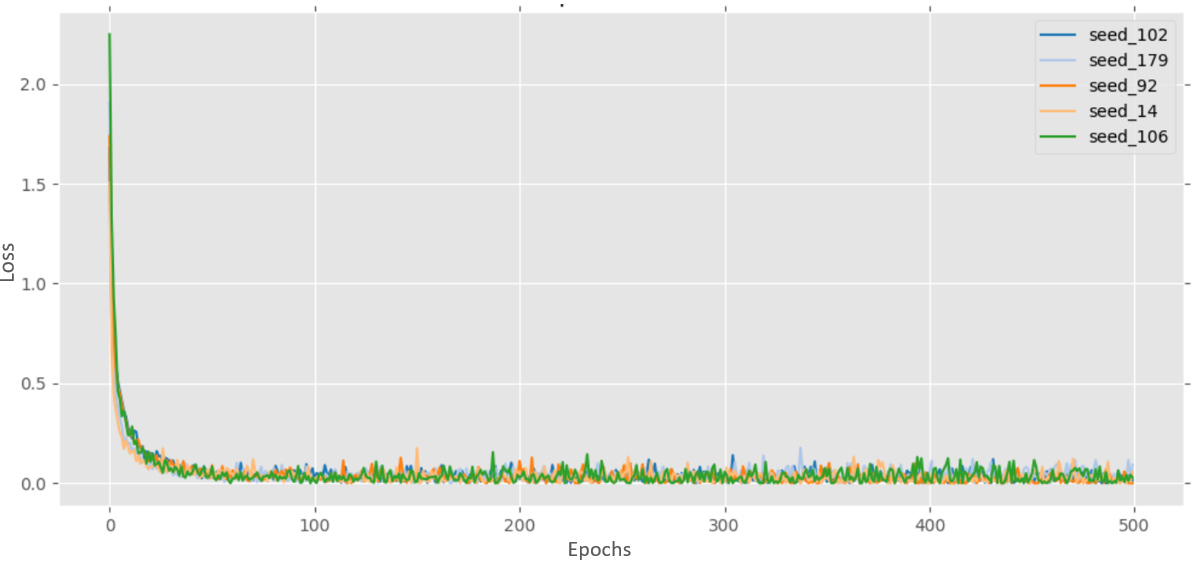} }}%
    \qquad
    \subfloat[Unsupervised Loss]{{\includegraphics[width=10cm]{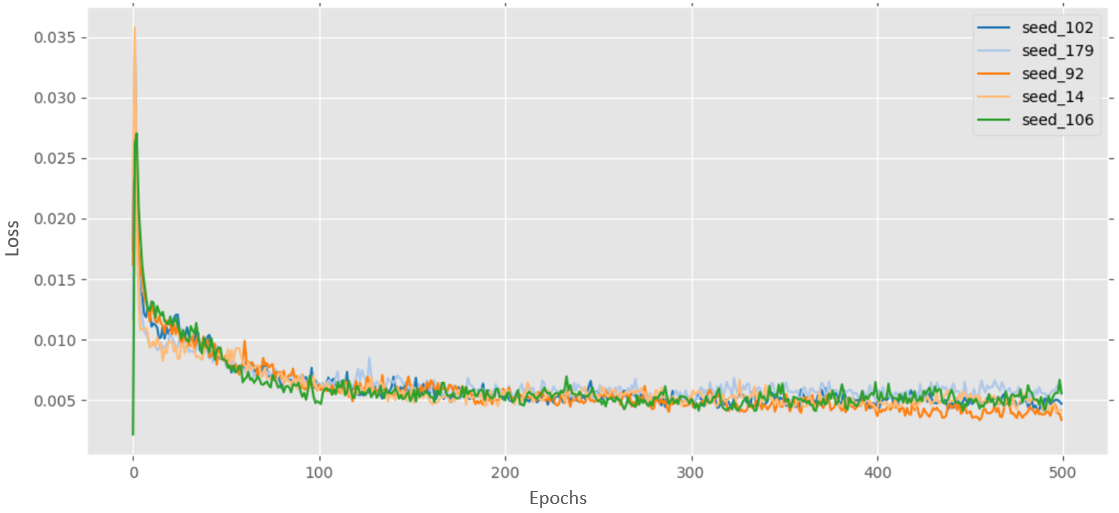} }}%
    \caption{Training loss behavior of MNIST using Temporal Ensembling with 500 Seeds}%
    \label{figmnist_seeds_500}%
\end{figure}

\begin{figure}[]
    \centering
    \subfloat[Supervised Loss]{{\includegraphics[width=10cm]{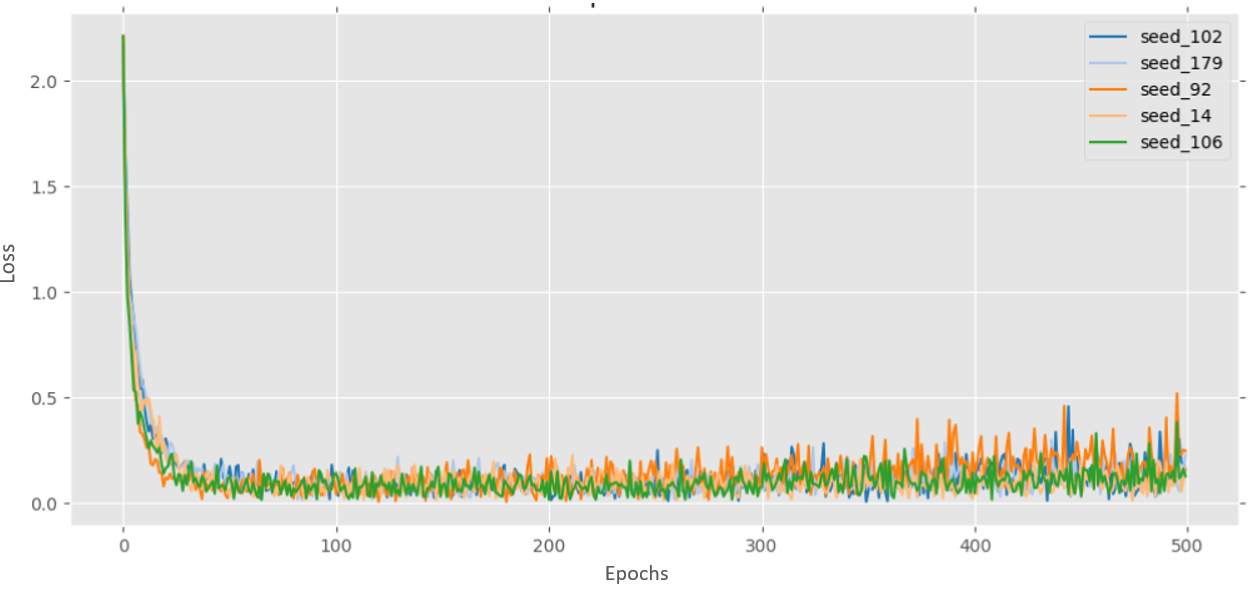} }}%
    \qquad
    \subfloat[Unsupervised Loss]{{\includegraphics[width=10cm]{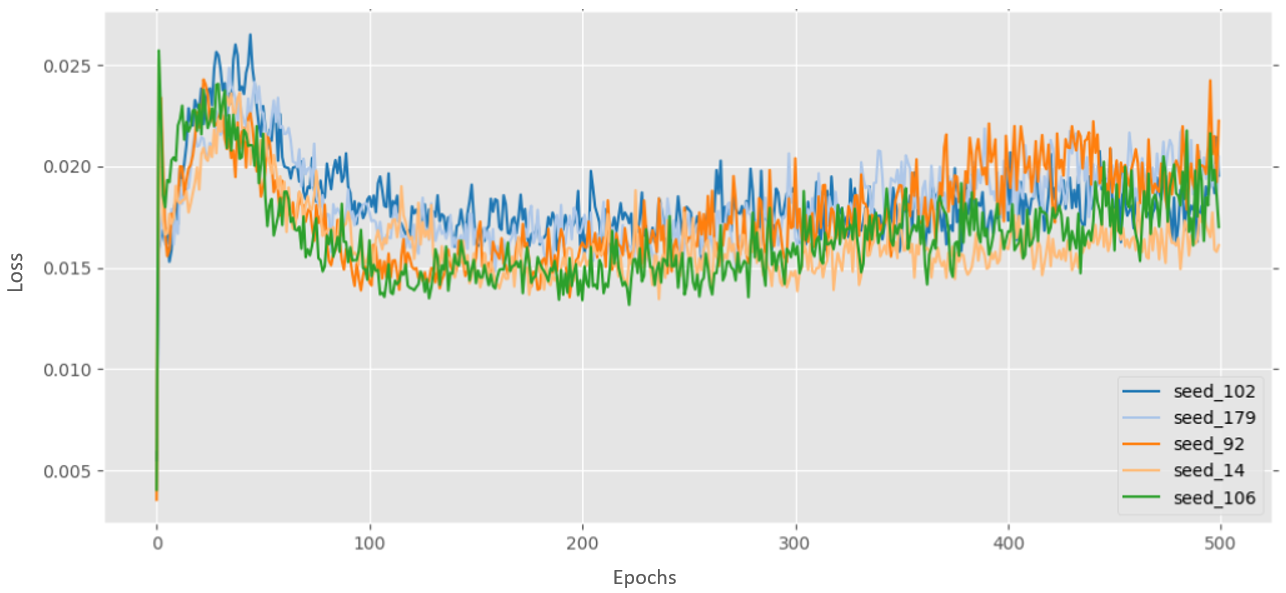} }}%
    \caption{Training loss behavior of KMNIST using Temporal Ensembling with 300 Seeds}%
    \label{figkmnist_seeds_300}%
\end{figure}

\begin{figure}[]
    \centering
    \subfloat[Supervised Loss]{{\includegraphics[width=10cm]{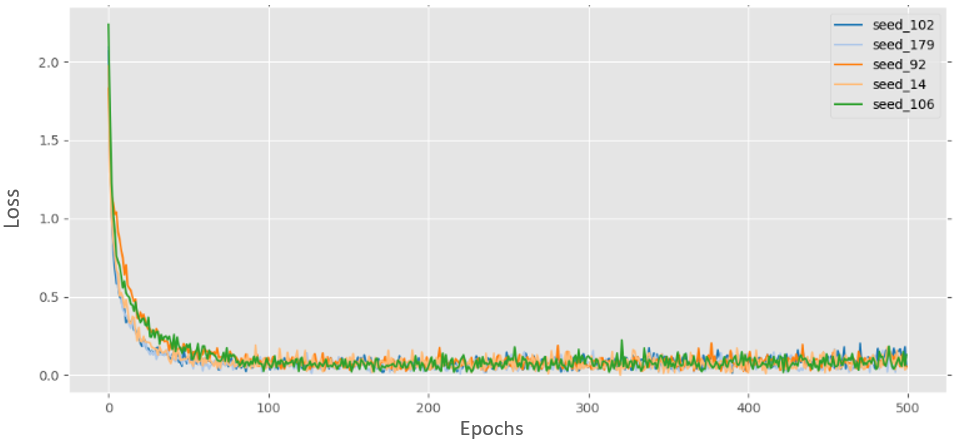} }}%
    \qquad
    \subfloat[Unsupervised Loss]{{\includegraphics[width=10cm]{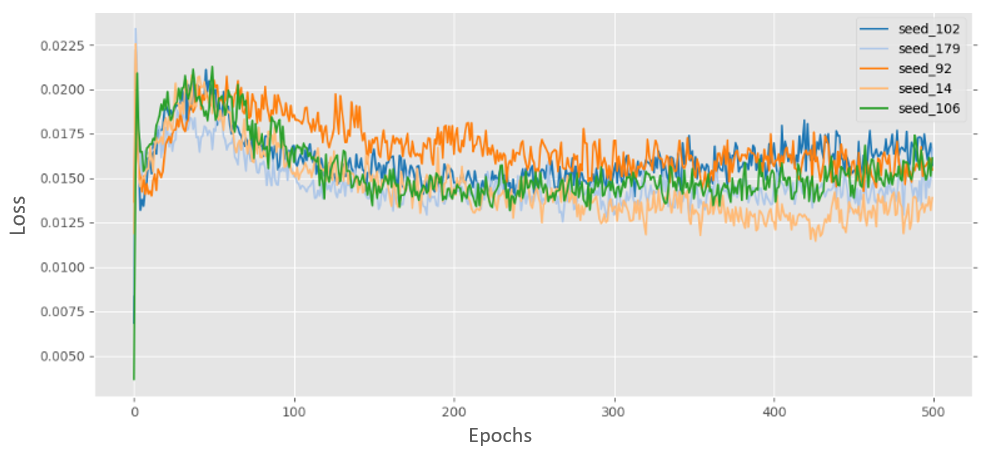} }}%
    \caption{Training loss behavior of KMNIST using Temporal Ensembling with 500 Seeds}%
    \label{figkmnist_seeds_500}%
\end{figure}

\begin{figure}[!htb]
\centering
{
    
    \subfloat[Examples from \textit{Seed\_14} used in 300 seed experiments]{{\includegraphics[width=10cm]{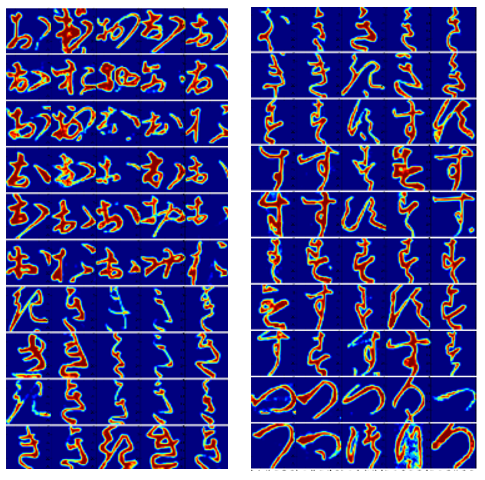} }}%
    \qquad
    \subfloat[Examples from \textit{Seed\_14} used in 500 seed experiments]{{\includegraphics[width=10cm]{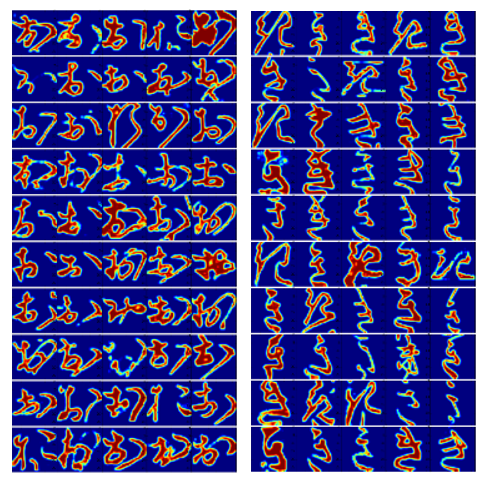} }}}%
    \caption{Examples KMNIST seeds used in experiments.}%
    \label{figkmnist_seeds_500_actual}%
\end{figure}

\begin{figure}[!htb]
    \centering
    \subfloat[Supervised Loss]{{\includegraphics[width=10cm]{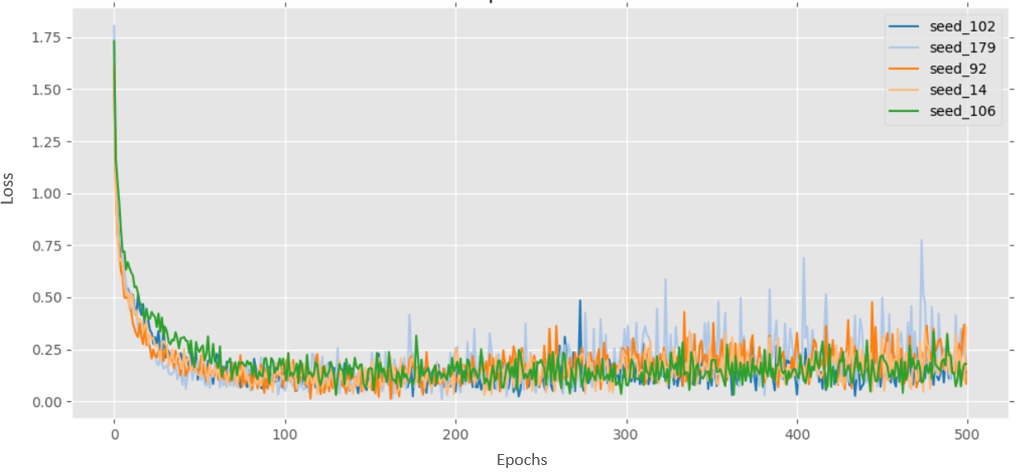} }}%
    \qquad
    \subfloat[Unsupervised Loss]{{\includegraphics[width=10cm]{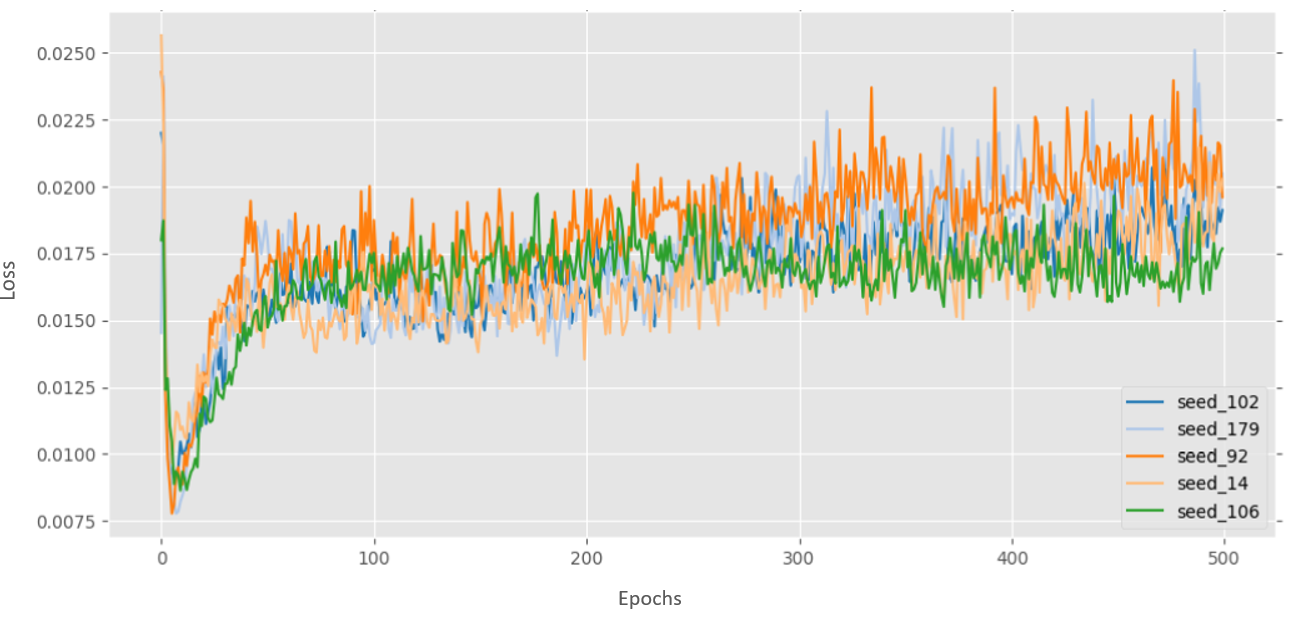} }}%
    \caption{Training loss behavior of Fashion-MNIST using Temporal Ensembling with 300 Seeds}%
    \label{figfashionmnist_seeds_300}%
\end{figure}

\begin{figure}[!htb]
    \centering
    \subfloat[Supervised Loss]{{\includegraphics[width=10cm]{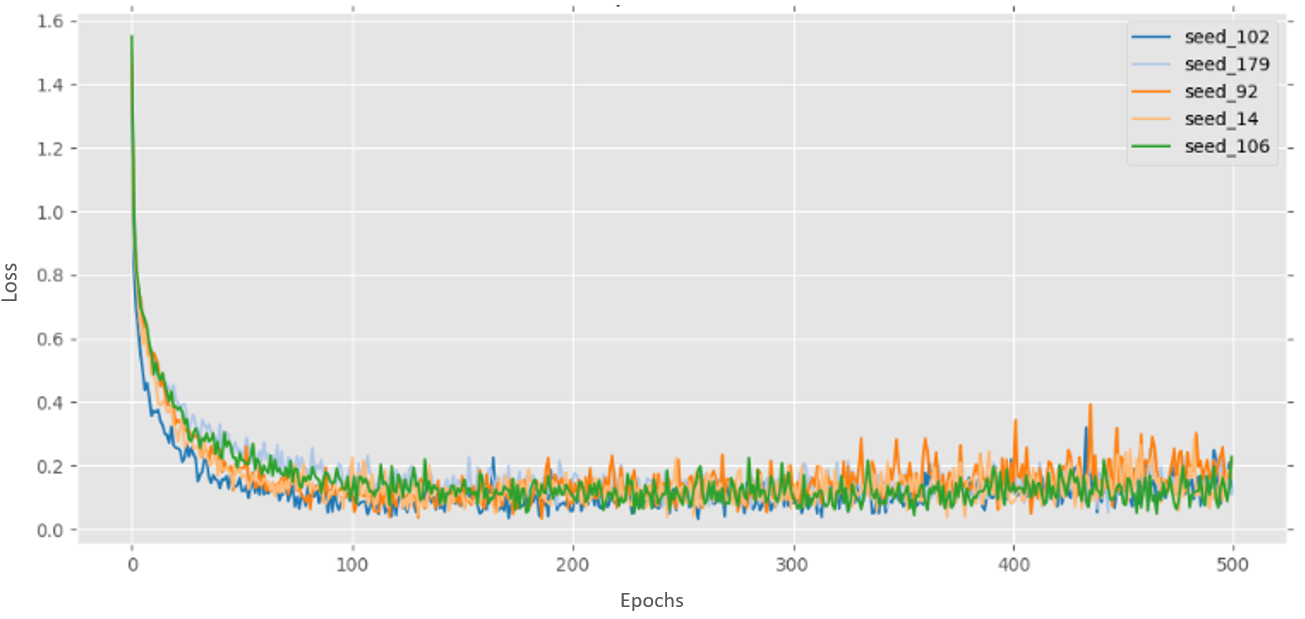} }}%
    \qquad
    \subfloat[Unsupervised Loss]{{\includegraphics[width=10cm]{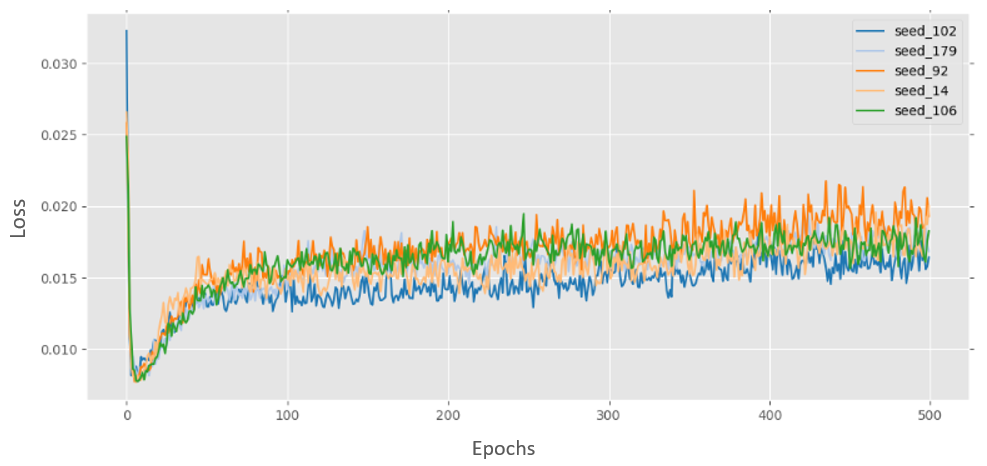} }}%
    \caption{Training loss behavior of Fashion-MNIST using Temporal Ensembling with 500 Seeds}%
    \label{figfashionmnist_seeds_500}%
\end{figure}

Accuracy score with standard deviation over five episodes of training each is shown in Tables \ref{table2_1} and \ref{table2_2}. Detailed experimental results are in section \ref{rq2appendix}. From Tables \ref{table2_1} and \ref{table2_2}, we can see that for MNIST datasets, the number of seeds indeed has no impact on the overall results with minor changes across varying sizes when tested at both 300 and 500 epochs, respectively. In fact, majority of the results are around 97\% with a maximum of 97.93\% when trained for 500 epochs, 400 labeled samples (0.5\% higher than highest results from Table 1) and a minimum of 96.93\% with 300 epochs, 200 labeled samples (0.5\% lower than best results from Table 1). 

Again this could be argued due to the nature of the dataset, where there is less non-uniformity (lower intraclass variability) within the classes, and train-test splits, and adding more seeds has minimal impact on the network learning behavior. Figure \ref{figmnist_seeds_300} and \ref{figmnist_seeds_500} also show loss curves for seed sizes of 300 and 500, trained for 500 epochs. As we can see, even with higher seeds, the convergence behavior seems very similar to RQ1 irrespective of training epochs. Further, this moderately supports the hypothesis that the number of seeds is a vital hyperparameter, and the performance of temporal ensembling indeed could be controlled by selecting an optimal amount of seeds. This also begs the question of rather than selecting the number of seeds, is it possible to select seeds that favor the test data distribution, which we shall revisit in section \ref{5.3}.

\begin{figure}[!htb]
\centering
{
    
    \subfloat[Examples from \textit{Seed\_14} used in 300 seed experiments]{{\includegraphics[width=10cm]{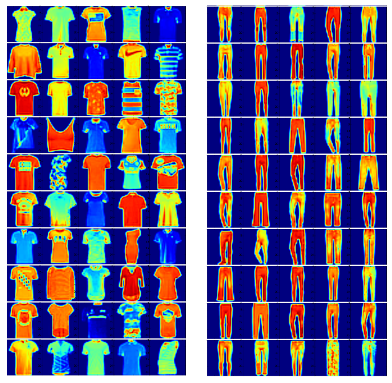} }}%
    \qquad
    \subfloat[Examples from \textit{Seed\_14} used in 500 seed experiments]{{\includegraphics[width=10cm]{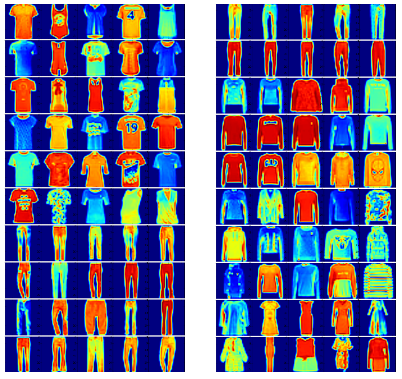} }}}%
    \caption{Examples Fashion-MNIST seeds used in experiments.}%
    \label{figfashionmnist_seeds_500_actual}%
\end{figure}

From Tables \ref{table2_1} and \ref{table2_2}, for KMNIST and Fashion-MNIST, we can see that results improve drastically with the addition of more seeds. To start with for KMNIST, the improvement in the number of seeds from 100 to 500 improves the result by 25\% with 300 epochs and 14.7\% with 500 epochs to result of 75\%. Meanwhile, Figure \ref{figkmnist_seeds_300} and \ref{figkmnist_seeds_500} show the loss curves with 300 and 500 seeds respectively across the five episodes of training under each epoch settings. As we can see with higher seeds, the network training behavior is indeed different, where we see lower values of both supervised and unsupervised loss. Further, such behavior is constant across various episodes of training. Examples from 300 and 500 seeds used during the preparation of KMNIST is as shown in Figure \ref{figkmnist_seeds_500_actual}. 

Fashion-MNIST again shows similar behavior like KMNIST except with net improvement of 8\%, with a maximum 81.53\% with 500 seeds. Yet the loss curves (Figures \ref{figfashionmnist_seeds_300} \& \ref{figfashionmnist_seeds_500}) and seeds (Figure \ref{figfashionmnist_seeds_500_actual}) show similar behavior like KMNIST.  What is very interesting to see is that performance of Fashion-MNIST is very close to the two-layer convolution baseline (TLCB) of 87\% (Table \ref{compare-res}), which is trained with complete 60k images. So even though temporal ensembling uses 150\% lesser images than TLCB, the performance difference is only 7\% lower. Similarly, for KMNIST, we can see that performance 17\% lower than the simplest K-Nearest Neighbor baseline of 92.1\% without any feature engineering (Table \ref{compare-res}).

\begin{table}[!htb]
\centering
\scalebox{0.9}{
\begin{tabular}{|c|c|c|c|}
\hline
\rowcolor[HTML]{CCCCCC} 
\textbf{Dataset} & \textbf{Approach} & \textbf{Seed Size} & \textbf{Results} \\ \hline
 & \textbf{K-Nearest Neighbors Baseline} & 60k & 92\% \\ \cline{2-4} 
\multirow{-2}{*}{\textbf{KMNIST}} & \textbf{Temporal Ensembling (500 epochs)} & 500 & 75.35\% \\ \hline
 & \textbf{Two-Layer Convolution Neural Network} & 60k & 87\% \\ \cline{2-4} 
\multirow{-2}{*}{\textbf{Fashion-MNIST}} & \textbf{Temporal Ensembling (500 epochs)} & 500 & 81.53\% \\ \hline
\end{tabular}}
\caption{Comparison of Results from Temporal Ensembling with Varying Seed Size (this work) against respective baselines for KMNIST and Fashion-MNIST with 60k images.}
\label{compare-res}
\end{table}
These two results confirm that there is some correlation between test data and seed size. Besides, this also shows that KMNIST to be a competitive baseline for temporal ensembling compared to MNIST and Fashion-MNIST in the context of intraclass variances. Also, these results strengthen our initial argument from section 1, that there is a need for experimental studies, such as those involving intraclass variability in semi-supervised learning approaches to find out what, apart from the number of labeled samples, what other aspects impacts performance. To summarise, our findings are:

\begin{itemize}
\itemsep0em
\item Seed size drastically impacts results under the conditions of intraclass variability, wherewith larger seeds size produces better results.
\item KMNIST and Fashion-MNIST show an improvement of 14.7\% and 8\% respectively with an increase in seed size from 100 to 500, producing competitive results again baselines trained on complete 60k images. While the net improvement obtained vary across both the datasets, the trend is consistent.
\item MNIST is a simple dataset, and from results, we can see that its not a strong baseline for temporal ensembling under the context of intraclass variability, which in turn could be fulfilled by KMNIST.
\item Besides, from results, we argue that there is a need for more detailed studies on the effect of intraclass variability in the context of general semi-supervised learning.
\end{itemize}

\subsection{RQ3: Effect of Seed Type on Temporal Ensembling under Intraclass Variability}\label{5.3}

Previously in section \ref{5.2}, we saw that with increasing seed size, the accuracy improved for all the datasets, and in the case of KMNIST and Fashion-MNIST, we observed substantial gains. Here we will examine how the type of seed will impact the results. This is similar to the argument of generalization in neural networks, where the trained model should at least correctly classify images identical to the one it was prepared for. However, in the context of temporal ensembling, the labeled examples serve two purposes, namely improving the quality of self-ensembled labels and, at the same time, also improve accuracy.

\begin{table}[!htb]
\centering
\scalebox{1.0}{
\begin{tabular}{|c|c|c|c|}
\hline
\rowcolor[HTML]{CCCCCC}
\textbf{Training   Episode} & \textbf{MNIST} & \textbf{KMNIST} & \textbf{Fashion-MNIST} \\ \hline
\textbf{\#1}                & 96.95\%        & 72.28\%         & 77.94\%               \\ \hline
\textbf{\#2}                & 97.03\%        & 77.40\%         & 80.05\%               \\ \hline
\textbf{\#3}                & 98.24\%        & 60.79\%         & 78.13\%               \\ \hline
\textbf{\#4}                & 98.58\%        & 74.69\%         & 78.15\%               \\ \hline
\textbf{\#5}                & 98.09\%        & 74.06\%         & 79.62\%               \\ \hline
\textbf{\#6}                & 98.14\%        & 74.58\%         & 79.25\%               \\ \hline
\textbf{\#7}                & 98.30\%        & 68.70\%         & 80.01\%               \\ \hline
\textbf{\#8}                & 97.64\%        & 67.30\%         & 76.87\%               \\ \hline
\textbf{\#9}                & 97.96\%        & 66.86\%         & 79.98\%               \\ \hline
\end{tabular}}
\caption{Results of MNIST, KMNIST and Fashion-MNIST trained for 500 epochs, 300 seeds across 10 episodes of training with different seed samples.}
\label{tab:my-table}
\end{table}

\begin{figure}[!htb]
    \centering
    \subfloat[Supervised Loss]{{\includegraphics[width=10cm]{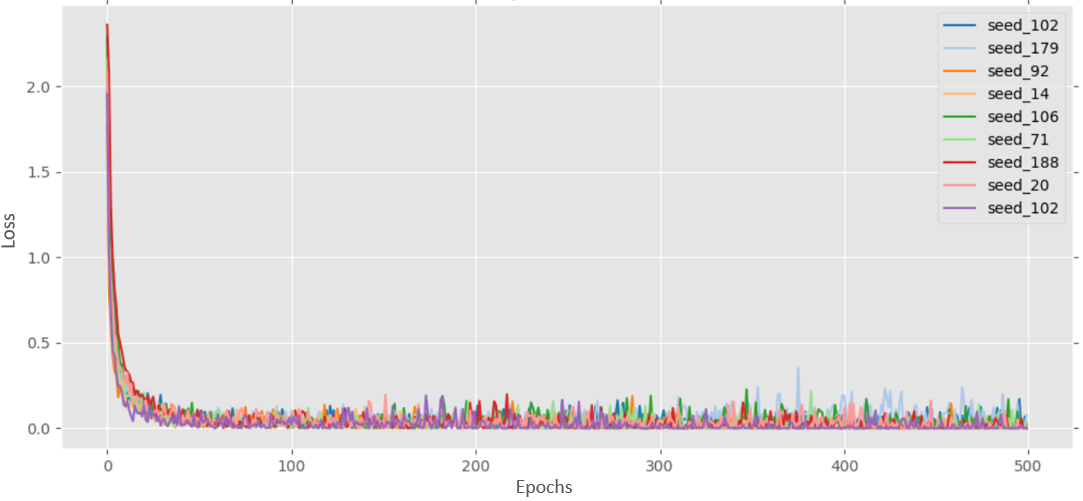} }}%
    \qquad
    \subfloat[Unsupervised Loss]{{\includegraphics[width=10cm]{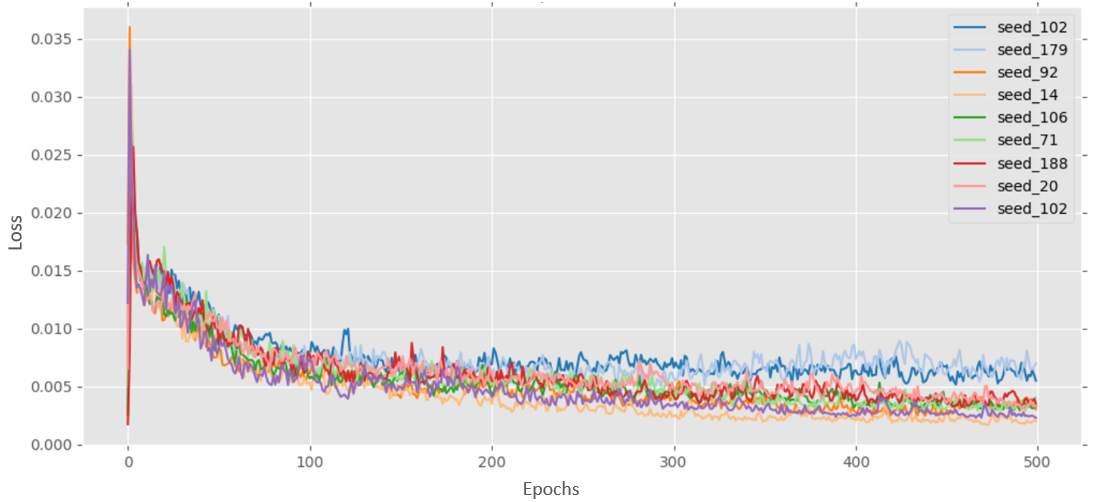} }}%
    \caption{Training loss behavior of MNIST trained for 10 episodes with varying seeds. Highest and Lowest models corresponds to seeds \textit{seed\_14} \& \textit{seed\_102} respectively}%
    \label{figmnist_seeds_500_rq3}%
\end{figure}

\begin{figure}[!htb]
    \centering
    \subfloat[Supervised Loss]{{\includegraphics[width=10cm]{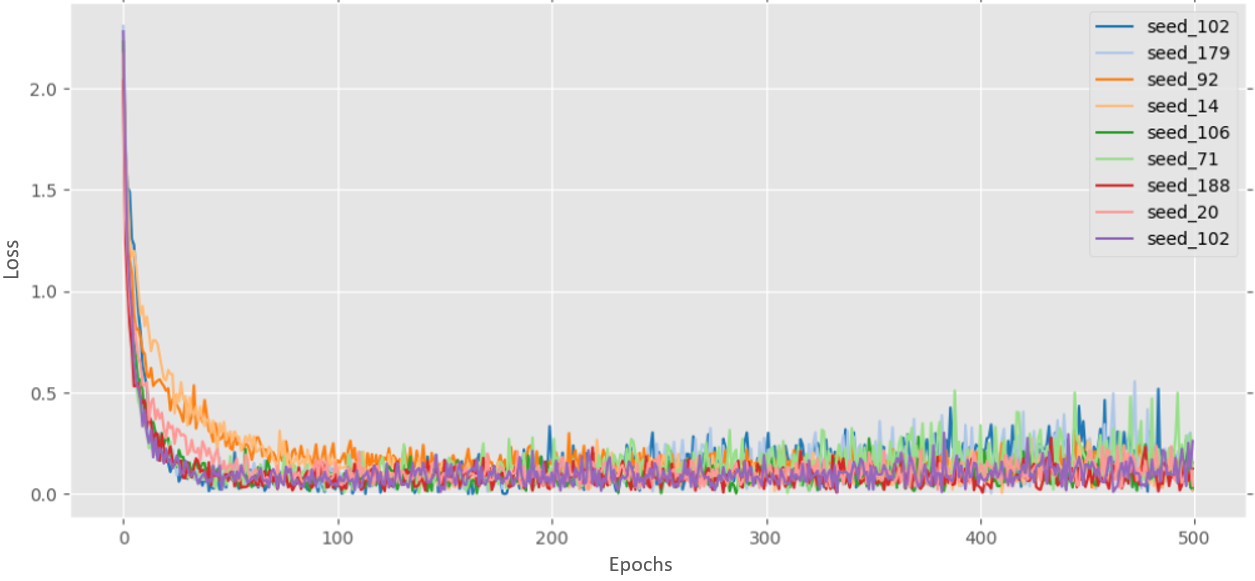} }}%
    \qquad
    \subfloat[Unsupervised Loss]{{\includegraphics[width=10cm]{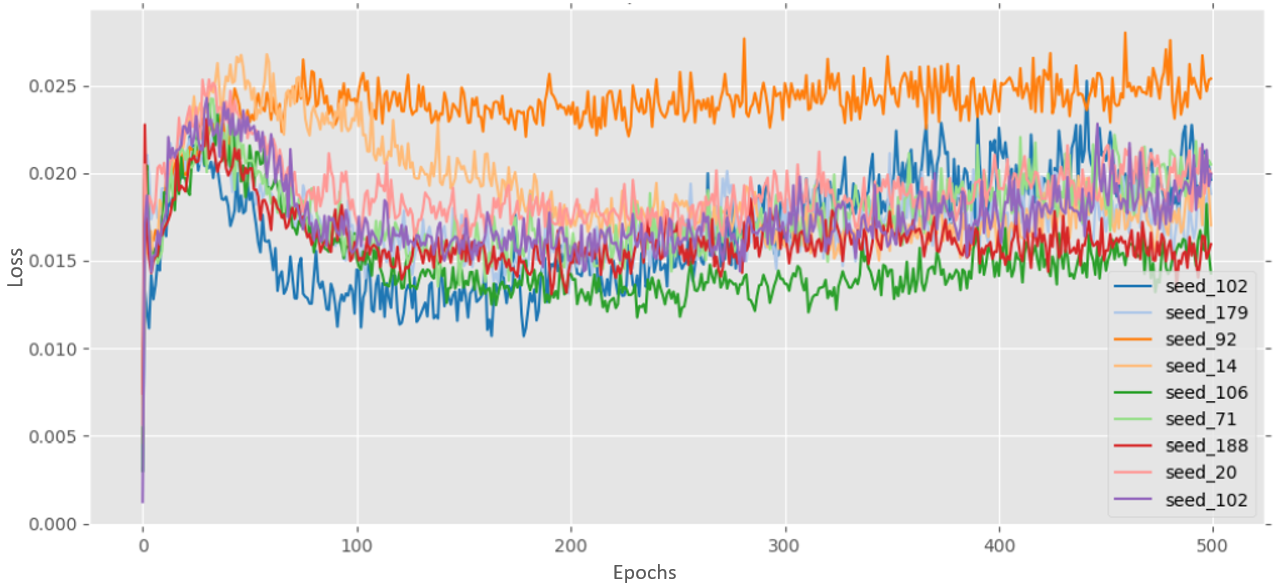} }}%
    \caption{Training loss behavior of KMNIST trained for 10 episodes with varying seeds. Highest and Lowest models corresponds to seeds \textit{seed\_179} \& \textit{seed\_92} respectively. See Figures \ref{mnistrq3_img} \& \ref{mnistrq3_img2} in appendix.}%
    \label{figkmnist_seeds_500_rq3}%
\end{figure}

\begin{figure}[!htb]
    \centering
    \subfloat[Supervised Loss]{{\includegraphics[width=10cm]{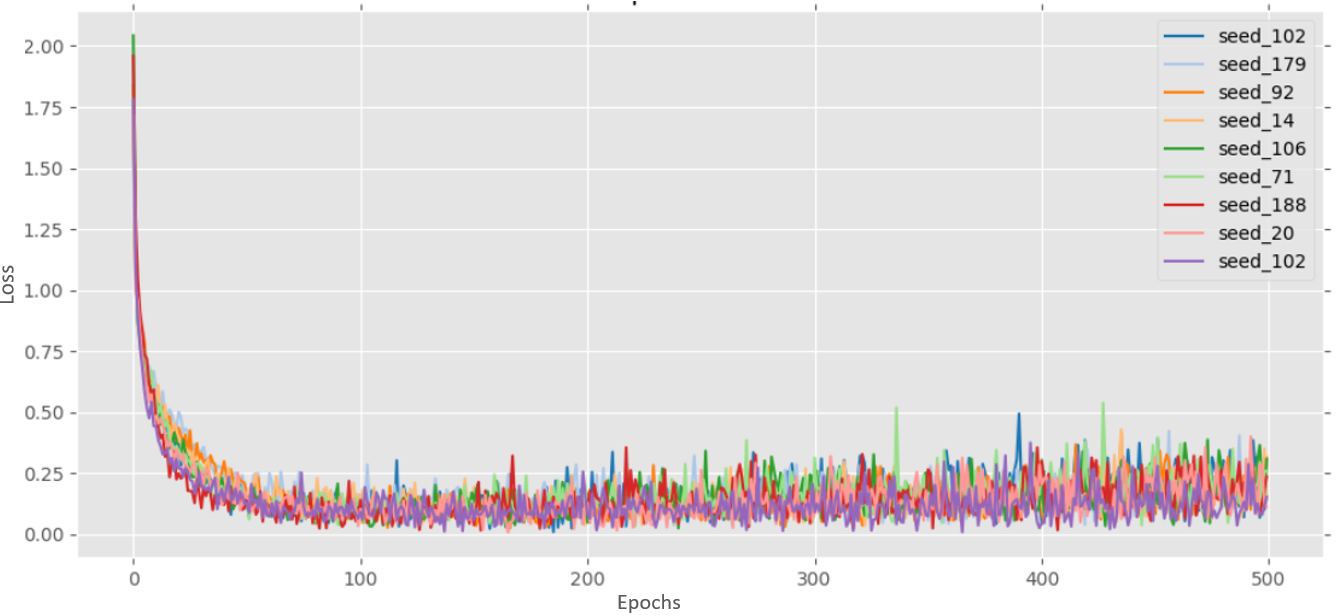} }}%
    \qquad
    \subfloat[Unsupervised Loss]{{\includegraphics[width=10cm]{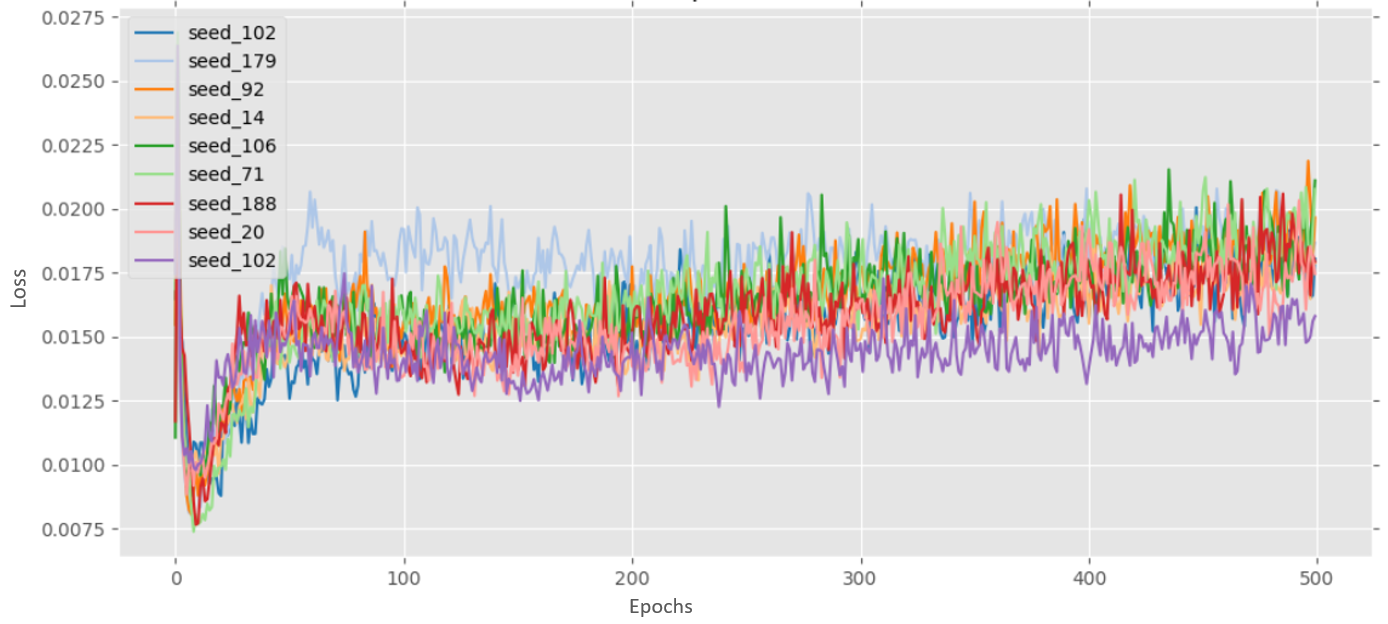} }}%
    \caption{Loss Behavior of KMNIST trained for 10 episodes with varying seeds. Highest and Lowest models corresponds to seeds \textit{seed\_102} \& \textit{seed\_20} respectively}%
    \label{figfashionmnist_seeds_500_rq3}%
\end{figure}

As such, one can hypothesize that starting with seeds with distributions closer to test set samples may achieve higher results and vice versa. Previous works have shown that all the datasets from section \ref{datasets} have examples of those similar to that of train distribution, and some are profoundly different. As such, to answer this RQ, we repeat experiments in line with RQ2, except we perform training with ten episodes with 300 seeds and 500 epochs. Also, to account for training randomness on the accuracy, each of the models was tested at every epoch. Finally, the model from the $500^{th}$ epoch was chosen due to its highest accuracy\footnote{Please note that, the results for RQ3 is still preliminary and requires further analysis \& validation.}. Consolidated results with standard deviation over ten episodes are as shown in Table \ref{tab:my-table}.

Comparing Table \ref{tab:my-table} with results from RQ2 (Table \ref{table2_2}),  with different samples of seeds, we can see that we have both higher and lower results than those from RQ2. Similar to findings from RQ1 and RQ2, MNIST produces results closer to 97\% despite random sampling of seeds across episodes. Further, the lowest effect of 96.95\% and the highest of 98.03\% could be observed. Figure \ref{figfashionmnist_seeds_500_rq3} shows, convergence behavior of both the models. As we can see, both the models converge similarly and inline with RQ1 and RQ2. 

Meanwhile, KMNIST and Fashion-MNIST show widely varying results (See Table \ref{tab:my-table}). To begin with, KMNIST produces the lowest accuracy of 66.86\% and the highest of 77.40\%. The lowest accuracy of KMNIST is closer to results obtained with 100 seeds, which in turn suggests that some of the seeds selected were practically useless. Similar behavior could be observed with the Fashion-MNIST dataset with the highest accuracy of 80.05 and lowest of 77.94\%, where again, the lowest result is close to that of results with 200 seeds in Table \ref{table2_2}.

The fact that the performance of some experiments with varying seeds is significantly lower than the other would raise a question that the models may have to overfit, and results obtained could be lower. However, as mentioned earlier, this was avoided by checking for model accuracy at each epoch. Also, if this is indeed the cases, then it, in turn, raises a new question of if there are samples that would allow faster and better convergence. Overall from these preliminary experiments 

\begin{itemize}
\itemsep0em
\item We find that seed selection indeed impacts the results across all the datasets. For the cases of KMNIST, we can see that the highest and lowest results obtained with different seeds differ by 2\% and 9\%, respectively. Similarly, for Fashion-MNIST, the numbers are 0.5\% and 3\%, respectively.
\item Performance with some of the seeds is significantly lower, which suggests some of the seeds offer more to training than others, which points to the identification of optimal seed images for temporal ensembling.
\item We further emphasize that the results shown here require further in-depth experimentation, analysis of results, and theoretical grounding$^1$.
\end{itemize}


\section{Conclusion and Future Work}\label{futurework}
This paper investigated the effect of intraclass variability on temporal ensembling. Firstly by analyzing different datasets, we showed that the dataset differs widely in terms of their intraclass variations. KMNIST offers the highest intraclass variability, followed by Fashion-MNIST and MNIST. From our study on RQ1, we find that given constant parameter settings, intraclass variability indeed affects the overall performance with KMNIST producing 60.66\% and Fashion-MNIST showing 71.31\%. Further, we also found that KMNIST and Fashion-MNIST present with a problem of lack of convergence of unsupervised loss and instead show an increase at higher epochs. Overall, our study of RQ1 suggests that temporal ensembling is not directly usable for datasets with high intraclass variability. 

From our review of the effect of seed size in RQ2, we see that higher seed size results in better accuracy across all the datasets KMNIST and Fashion-MNIST show an improvement of 14.7\% and 8\% respectively with an increase in seed size from 100 to 500, producing competitive results again baselines trained on complete 60k images. Further, we could see that KMNIST serves as a competitive baseline for temporal ensembling as it accounts for intraclass variability. Finally, in RQ3, we demonstrate that with different seed images, we get different results. 

Overall across a broad range of datasets, we examined how intraclass variability temporal ensembling performs. While there is considerable variation within the classes across the datasets, it is consistent
that the class with more intraclass variability is harder to do classify. This
connection with intraclass variability occurs to such an extent that, in fact, temporal ensembling with 1/3 of the selected seeds offers similar accuracy (See section \ref{5.3}). However, at the same time, this result serves as a decent baseline. We believe seeds are critical, and the temporal ensembling with different seeds is not sufficiently effective at generalizing beyond the image contexts found in training data. To close this gap and advance temporal ensembling for practical applications various aspects need more in-depth exploration including 
(a) effect of varying other hyperparameters (Table \ref{param}) in temporal ensembling, (b) relationship between seed types and data distribution and (c) reason for the rise in the unsupervised loss at higher epochs.

\bibliographystyle{coling}
\bibliography{coling2020}

\section*{Acknowledgements}
We would like to thank Johan Ferret \cite{ferret_2018}, for his pytorch 0.4 implementation of temporal ensembling.

\section*{Author Contributions}
MR designed the study. SV carried out the primary experiments. Both the authors carried out analysis, secondary experiments, contributed to developing the outline and editing the manuscript. MR wrote the manuscript and SV reviewed the manuscript.


\section{Appendix}
We now present various experimental details and results for each of the RQ\'s.

\subsection{Detailed Experimental Results of RQ1}\label{rq1appendix}
Tables \ref{tablerq1exps1.1}-\ref{tablerq1exps1.3} present detailed results for RQ1.

\begin{table}[!ht]
\centering
\scalebox{1.0}{
\begin{tabular}{|c|c|c|c|c|}
\hline
\textbf{Dataset}                         & \textbf{Epochs}               & \textbf{Experiment} & \textbf{Accuracy} & \textbf{Accuracy (Best Model)} \\ \hline
\multirow{18}{*}{\textbf{MNIST}}         & \multirow{6}{*}{\textbf{100}} & 1                   & 95.78\%           & 95.78\%                        \\ \cline{3-5} 
                                         &                               & 2                   & 93.86\%           & 91.37\%                        \\ \cline{3-5} 
                                         &                               & 3                   & 95.23\%           & 90.59\%                        \\ \cline{3-5} 
                                         &                               & 4                   & 95.04\%           & 93.11\%                        \\ \cline{3-5} 
                                         &                               & 5                   & 95.66\%           & 95.51\%                        \\ \cline{3-5} 
                                         &                               & \textbf{Average}    & 95.11\%           & 93.27\%                        \\ \cline{2-5} 
                                         & \multirow{6}{*}{\textbf{300}} & 1                   & 95.89\%           & 95.70\%                        \\ \cline{3-5} 
                                         &                               & 2                   & 96.98\%           & 97.13\%                        \\ \cline{3-5} 
                                         &                               & 3                   & 97.86\%           & 98.05\%                        \\ \cline{3-5} 
                                         &                               & 4                   & 97.47\%           & 97.56\%                        \\ \cline{3-5} 
                                         &                               & 5                   & 97.45\%           & 97.60\%                        \\ \cline{3-5} 
                                         &                               & \textbf{Average}    & 97.13\%           & 97.21\%                        \\ \cline{2-5} 
                                         & \multirow{6}{*}{\textbf{500}} & 1                   & 98.15\%           & 98.16\%                        \\ \cline{3-5} 
                                         &                               & 2                   & 95.58\%           & 95.33\%                        \\ \cline{3-5} 
                                         &                               & 3                   & 97.59\%           & 97.61\%                        \\ \cline{3-5} 
                                         &                               & 4                   & 98.02\%           & 98.18\%                        \\ \cline{3-5} 
                                         &                               & 5                   & 97.78\%           & 97.86\%                        \\ \cline{3-5} 
                                         &                               & Average             & 97.42\%           & 97.43\%                        \\ \hline
\end{tabular}}
\caption{Detailed Experimental results for RQ1 on MNIST}
\label{tablerq1exps1.1}
\end{table}

\begin{table}[!ht]
\centering
\scalebox{1.0}{
\begin{tabular}{|c|c|c|c|c|}
\hline
\textbf{Dataset}                         & \textbf{Epochs}               & \textbf{Experiment} & \textbf{Accuracy} & \textbf{Accuracy (Best Model)} \\ \hline
\multirow{18}{*}{\textbf{KMNIST}}        & \multirow{6}{*}{\textbf{100}} & 1                   & 59.20\%           & 58.64\%                        \\ \cline{3-5} 
                                         &                               & 2                   & 58.27\%           & 58.55\%                        \\ \cline{3-5} 
                                         &                               & 3                   & 59.41\%           & 61.77\%                        \\ \cline{3-5} 
                                         &                               & 4                   & 10.00\%           & 9.99\%                         \\ \cline{3-5} 
                                         &                               & 5                   & 62.41\%           & 61.31\%                        \\ \cline{3-5} 
                                         &                               & \textbf{Average}    & 49.86\%           & 50.05\%                        \\ \cline{2-5} 
                                         & \multirow{6}{*}{\textbf{300}} & 1                   & 53.45\%           & 55.98\%                        \\ \cline{3-5} 
                                         &                               & 2                   & 10.00\%           & 10.00\%                        \\ \cline{3-5} 
                                         &                               & 3                   & 57.84\%           & 64.69\%                        \\ \cline{3-5} 
                                         &                               & 4                   & 57.09\%           & 61.97\%                        \\ \cline{3-5} 
                                         &                               & 5                   & 47.84\%           & 56.85\%                        \\ \cline{3-5} 
                                         &                               & \textbf{Average}    & 45.24\%           & 49.90\%                        \\ \cline{2-5} 
                                         & \multirow{6}{*}{\textbf{500}} & 1                   & 50.90\%           & 56.49\%                        \\ \cline{3-5} 
                                         &                               & 2                   & 50.18\%           & 60.35\%                        \\ \cline{3-5} 
                                         &                               & 3                   & 49.94\%           & 63.86\%                        \\ \cline{3-5} 
                                         &                               & 4                   & 50.39\%           & 55.26\%                        \\ \cline{3-5} 
                                         &                               & 5                   & 53.52\%           & 64.32\%                        \\ \cline{3-5} 
                                         &                               & \textbf{Average}    & 50.99\%           & 60.06\%                        \\ \hline
\end{tabular}}
\caption{Detailed Experimental results for RQ1 on KMNIST}
\label{tablerq1exps1.2}
\end{table}

\begin{table}[!ht]
\centering
\scalebox{1.0}{
\begin{tabular}{|c|c|c|c|c|}
\hline
\textbf{Dataset}                         & \textbf{Epochs}               & \textbf{Experiment} & \textbf{Accuracy} & \textbf{Accuracy (Best Model)} \\ \hline
\multirow{18}{*}{\textbf{Fashion-MNIST}} & \multirow{6}{*}{\textbf{100}} & 1                   & 75.94\%           & 75.28\%                        \\ \cline{3-5} 
                                         &                               & 2                   & 72.82\%           & 73.00\%                        \\ \cline{3-5} 
                                         &                               & 3                   & 72.31\%           & 73.28\%                        \\ \cline{3-5} 
                                         &                               & 4                   & 73.74\%           & 72.48\%                        \\ \cline{3-5} 
                                         &                               & 5                   & 75.55\%           & 76.67\%                        \\ \cline{3-5} 
                                         &                               & Average             & 74.07\%           & 74.14\%                        \\ \cline{2-5} 
                                         & \multirow{6}{*}{\textbf{300}} & 1                   & 68.61\%           & 73.14\%                        \\ \cline{3-5} 
                                         &                               & 2                   & 73.02\%           & 74.47\%                        \\ \cline{3-5} 
                                         &                               & 3                   & 69.86\%           & 73.43\%                        \\ \cline{3-5} 
                                         &                               & 4                   & 74.49\%           & 75.63\%                        \\ \cline{3-5} 
                                         &                               & 5                   & 71.31\%           & 73.09\%                        \\ \cline{3-5} 
                                         &                               & \textbf{Average}    & 71.46\%           & 73.95\%                        \\ \cline{2-5} 
                                         & \multirow{6}{*}{\textbf{500}} & 1                   & 70.69\%           & 72.46\%                        \\ \cline{3-5} 
                                         &                               & 2                   & 66.58\%           & 69.86\%                        \\ \cline{3-5} 
                                         &                               & 3                   & 66.99\%           & 73.41\%                        \\ \cline{3-5} 
                                         &                               & 4                   & 65.74\%           & 67.21\%                        \\ \cline{3-5} 
                                         &                               & 5                   & 70.08\%           & 73.59\%                        \\ \cline{3-5} 
                                         &                               & \textbf{Average}    & 68.02\%           & 71.31\%                        \\ \hline
\end{tabular}}
\caption{Detailed Experimental results for RQ1 on Fashion-MNIST}
\label{tablerq1exps1.3}
\end{table}

\subsection{Detailed Experimental Results of RQ2}\label{rq2appendix}
Tables \ref{200seeds}-\ref{500seeds}, shows experimental results obtained with varying seeds from 200 to 500 respectively.
\begin{table}[!htb]
\centering
\scalebox{0.8}{
\begin{tabular}{|c|c|c|c|c|}
\hline
\textbf{Dataset} & \textbf{Epochs} & \textbf{Experiment} & \textbf{Accuracy} & \textbf{Accuracy (Best Model)} \\ \hline
\multirow{18}{*}{\textbf{MNIST}} & \multirow{6}{*}{\textbf{100}} & 1 & 97.57\% & 97.48\% \\ \cline{3-5} 
 &  & 2 & 95.50\% & 95.67\% \\ \cline{3-5} 
 &  & 3 & 96.77\% & 96.77\% \\ \cline{3-5} 
 &  & 4 & 96.64\% & 96.51\% \\ \cline{3-5} 
 &  & 5 & 95.40\% & 94.85\% \\ \cline{3-5} 
 &  & Average & 96.38\% & 96.26\% \\ \cline{2-5} 
 & \multirow{6}{*}{\textbf{300}} & 1 & 97.59\% & 97.23\% \\ \cline{3-5} 
 &  & 2 & 98.09\% & 98.14\% \\ \cline{3-5} 
 &  & 3 & 97.93\% & 98.00\% \\ \cline{3-5} 
 &  & 4 & 93.86\% & 93.75\% \\ \cline{3-5} 
 &  & 5 & 97.16\% & 97.53\% \\ \cline{3-5} 
 &  & Average & 96.93\% & 96.93\% \\ \cline{2-5} 
 & \multirow{6}{*}{\textbf{500}} & 1 & 97.63\% & 97.63\% \\ \cline{3-5} 
 &  & 2 & 97.12\% & 96.90\% \\ \cline{3-5} 
 &  & 3 & 97.74\% & 98.24\% \\ \cline{3-5} 
 &  & 4 & 95.70\% & 96.42\% \\ \cline{3-5} 
 &  & 5 & 97.75\% & 98.46\% \\ \cline{3-5} 
 &  & Average & 97.19\% & 97.53\% \\ \hline
\multirow{18}{*}{\textbf{KMNIST}} & \multirow{6}{*}{\textbf{100}} & 1 & 63.97\% & 64.32\% \\ \cline{3-5} 
 &  & 2 & 69.24\% & 69.44\% \\ \cline{3-5} 
 &  & 3 & 66.85\% & 64.85\% \\ \cline{3-5} 
 &  & 4 & 61.56\% & 58.81\% \\ \cline{3-5} 
 &  & 5 & 68.41\% & 68.21\% \\ \cline{3-5} 
 &  & Average & 66.01\% & 65.13\% \\ \cline{2-5} 
 & \multirow{6}{*}{\textbf{300}} & 1 & 62.87\% & 62.73\% \\ \cline{3-5} 
 &  & 2 & 64.74\% & 62.75\% \\ \cline{3-5} 
 &  & 3 & 68.03\% & 68.99\% \\ \cline{3-5} 
 &  & 4 & 67.29\% & 68.00\% \\ \cline{3-5} 
 &  & 5 & 70.54\% & 71.50\% \\ \cline{3-5} 
 &  & Average & 66.69\% & 66.79\% \\ \cline{2-5} 
 & \multirow{6}{*}{\textbf{500}} & 1 & 65.17\% & 70.11\% \\ \cline{3-5} 
 &  & 2 & 62.11\% & 70.78\% \\ \cline{3-5} 
 &  & 3 & 65.23\% & 67.17\% \\ \cline{3-5} 
 &  & 4 & 66.75\% & 70.91\% \\ \cline{3-5} 
 &  & 5 & 70.57\% & 71.00\% \\ \cline{3-5} 
 &  & Average & 65.97\% & 69.99\% \\ \hline
\multirow{18}{*}{\textbf{Fashion-MNIST}} & \multirow{6}{*}{\textbf{100}} & 1 & 77.14\% & 77.19\% \\ \cline{3-5} 
 &  & 2 & 78.71\% & 77.80\% \\ \cline{3-5} 
 &  & 3 & 74.22\% & 72.40\% \\ \cline{3-5} 
 &  & 4 & 77.11\% & 76.55\% \\ \cline{3-5} 
 &  & 5 & 77.39\% & 76.21\% \\ \cline{3-5} 
 &  & Average & 76.91\% & 76.03\% \\ \cline{2-5} 
 & \multirow{6}{*}{\textbf{300}} & 1 & 75.01\% & 74.93\% \\ \cline{3-5} 
 &  & 2 & 77.02\% & 77.57\% \\ \cline{3-5} 
 &  & 3 & 74.26\% & 77.62\% \\ \cline{3-5} 
 &  & 4 & 74.88\% & 75.22\% \\ \cline{3-5} 
 &  & 5 & 78.14\% & 78.89\% \\ \cline{3-5} 
 &  & Average & 75.43\% & 76.71\% \\ \cline{2-5} 
 & \multirow{6}{*}{\textbf{500}} & 1 & 76.72\% & 78.19\% \\ \cline{3-5} 
 &  & 2 & 75.57\% & 76.05\% \\ \cline{3-5} 
 &  & 3 & 74.82\% & 75.87\% \\ \cline{3-5} 
 &  & 4 & 76.60\% & 77.10\% \\ \cline{3-5} 
 &  & 5 & 76.10\% & 76.86\% \\ \cline{3-5} 
 &  & Average & 75.96\% & 76.81\% \\ \hline
\end{tabular}}
\caption{Results of MNIST, KMNIST and Fashion-MNIST with 200 seeds.}
\label{200seeds}
\end{table}

\begin{table}[!htb]
\centering
\scalebox{0.8}{
\begin{tabular}{|c|c|c|c|c|}
\hline
\textbf{Dataset} & \textbf{Epochs} & \textbf{Experiment} & \textbf{Accuracy} & \textbf{Accuracy (Best Model)} \\ \hline
\multirow{18}{*}{\textbf{MNIST}} & \multirow{6}{*}{\textbf{100}} & 1 & 96.42\% & 96.71\% \\ \cline{3-5} 
 &  & 2 & 95.65\% & 95.63\% \\ \cline{3-5} 
 &  & 3 & 96.67\% & 96.45\% \\ \cline{3-5} 
 &  & 4 & 96.35\% & 96.67\% \\ \cline{3-5} 
 &  & 5 & 96.12\% & 95.95\% \\ \cline{3-5} 
 &  & \textbf{Average} & \textbf{96.24\%} & \textbf{96.28\%} \\ \cline{2-5} 
 & \multirow{6}{*}{\textbf{300}} & 1 & 97.85\% & 97.82\% \\ \cline{3-5} 
 &  & 2 & 97.96\% & 98.04\% \\ \cline{3-5} 
 &  & 3 & 97.91\% & 97.91\% \\ \cline{3-5} 
 &  & 4 & 97.45\% & 97.42\% \\ \cline{3-5} 
 &  & 5 & 98.06\% & 97.92\% \\ \cline{3-5} 
 &  & \textbf{Average} & \textbf{97.85\%} & \textbf{97.82\%} \\ \cline{2-5} 
 & \multirow{6}{*}{\textbf{500}} & 1 & 97.78\% & 97.85\% \\ \cline{3-5} 
 &  & 2 & 97.11\% & 97.70\% \\ \cline{3-5} 
 &  & 3 & 98.12\% & 98.12\% \\ \cline{3-5} 
 &  & 4 & 97.57\% & 97.65\% \\ \cline{3-5} 
 &  & 5 & 97.67\% & 97.25\% \\ \cline{3-5} 
 &  & \textbf{Average} & \textbf{97.65\%} & \textbf{97.71\%} \\ \hline
\multirow{18}{*}{\textbf{KMNIST}} & \multirow{6}{*}{\textbf{100}} & 1 & 73.91\% & 70.67\% \\ \cline{3-5} 
 &  & 2 & 68.88\% & 68.74\% \\ \cline{3-5} 
 &  & 3 & 69.00\% & 65.49\% \\ \cline{3-5} 
 &  & 4 & 67.53\% & 68.54\% \\ \cline{3-5} 
 &  & 5 & 71.60\% & 70.24\% \\ \cline{3-5} 
 &  & \textbf{Average} & \textbf{70.18\%} & \textbf{68.74\%} \\ \cline{2-5} 
 & \multirow{6}{*}{\textbf{300}} & 1 & 70.34\% & 71.18\% \\ \cline{3-5} 
 &  & 2 & 65.85\% & 66.05\% \\ \cline{3-5} 
 &  & 3 & 70.64\% & 71.04\% \\ \cline{3-5} 
 &  & 4 & 71.83\% & 71.09\% \\ \cline{3-5} 
 &  & 5 & 70.74\% & 70.75\% \\ \cline{3-5} 
 &  & \textbf{Average} & \textbf{69.88\%} & \textbf{70.02\%} \\ \cline{2-5} 
 & \multirow{6}{*}{\textbf{500}} & 1 & 68.80\% & 71.64\% \\ \cline{3-5} 
 &  & 2 & 67.87\% & 69.48\% \\ \cline{3-5} 
 &  & 3 & 63.09\% & 64.51\% \\ \cline{3-5} 
 &  & 4 & 69.39\% & 72.52\% \\ \cline{3-5} 
 &  & 5 & 70.17\% & 72.38\% \\ \cline{3-5} 
 &  & \textbf{Average} & \textbf{67.86\%} & \textbf{70.11\%} \\ \hline
\multirow{18}{*}{\textbf{Fashion-MNIST}} & \multirow{6}{*}{\textbf{100}} & 1 & 78.36\% & 78.32\% \\ \cline{3-5} 
 &  & 2 & 78.76\% & 78.47\% \\ \cline{3-5} 
 &  & 3 & 79.43\% & 79.20\% \\ \cline{3-5} 
 &  & 4 & 77.05\% & 77.77\% \\ \cline{3-5} 
 &  & 5 & 78.44\% & 78.94\% \\ \cline{3-5} 
 &  & \textbf{Average} & \textbf{78.41\%} & \textbf{78.54\%} \\ \cline{2-5} 
 & \multirow{6}{*}{\textbf{300}} & 1 & 77.29\% & 77.55\% \\ \cline{3-5} 
 &  & 2 & 77.89\% & 78.14\% \\ \cline{3-5} 
 &  & 3 & 78.23\% & 77.69\% \\ \cline{3-5} 
 &  & 4 & 80.26\% & 80.26\% \\ \cline{3-5} 
 &  & 5 & 79.37\% & 79.88\% \\ \cline{3-5} 
 &  & \textbf{Average} & \textbf{78.61\%} & \textbf{78.70\%} \\ \cline{2-5} 
 & \multirow{6}{*}{\textbf{500}} & 1 & 79.92\% & 79.69\% \\ \cline{3-5} 
 &  & 2 & 78.99\% & 81.14\% \\ \cline{3-5} 
 &  & 3 & 78.62\% & 79.69\% \\ \cline{3-5} 
 &  & 4 & 80.32\% & 80.65\% \\ \cline{3-5} 
 &  & 5 & 79.61\% & 78.44\% \\ \cline{3-5} 
 &  & \textbf{Average} & \textbf{79.49\%} & \textbf{79.92\%} \\ \hline
\end{tabular}}
\caption{Results of MNIST, KMNIST and Fashion-MNIST with 300 seeds.}
\label{300seeds}
\end{table}

\begin{table}[!htb]
\centering
\scalebox{0.8}{
\begin{tabular}{|c|c|c|c|c|}
\hline
\textbf{Dataset} & \textbf{Epochs} & \textbf{Experiment} & \textbf{Accuracy} & \textbf{Accuracy (Best Model)} \\ \hline
\multirow{18}{*}{\textbf{MNIST}} & \multirow{6}{*}{\textbf{100}} & 1 & 97.18\% & 96.62\% \\ \cline{3-5} 
 &  & 2 & 95.30\% & 96.19\% \\ \cline{3-5} 
 &  & 3 & 97.02\% & 95.18\% \\ \cline{3-5} 
 &  & 4 & 96.29\% & 96.24\% \\ \cline{3-5} 
 &  & 5 & 96.03\% & 95.90\% \\ \cline{3-5} 
 &  & \textbf{Average} & \textbf{96.36\%} & \textbf{96.03\%} \\ \cline{2-5} 
 & \multirow{6}{*}{\textbf{300}} & 1 & 97.23\% & 97.39\% \\ \cline{3-5} 
 &  & 2 & 97.22\% & 96.78\% \\ \cline{3-5} 
 &  & 3 & 97.45\% & 97.56\% \\ \cline{3-5} 
 &  & 4 & 97.77\% & 97.66\% \\ \cline{3-5} 
 &  & 5 & 97.46\% & 97.65\% \\ \cline{3-5} 
 &  & \textbf{Average} & \textbf{97.43\%} & \textbf{97.41\%} \\ \cline{2-5} 
 & \multirow{6}{*}{\textbf{500}} & 1 & 97.95\% & 97.86\% \\ \cline{3-5} 
 &  & 2 & 97.46\% & 97.68\% \\ \cline{3-5} 
 &  & 3 & 97.82\% & 97.70\% \\ \cline{3-5} 
 &  & 4 & 97.41\% & 97.55\% \\ \cline{3-5} 
 &  & 5 & 98.05\% & 97.85\% \\ \cline{3-5} 
 &  & \textbf{Average} & \textbf{97.74\%} & \textbf{97.73\%} \\ \hline
\multirow{18}{*}{\textbf{KMNIST}} & \multirow{6}{*}{\textbf{100}} & 1 & 71.05\% & 71.82\% \\ \cline{3-5} 
 &  & 2 & 71.67\% & 71.67\% \\ \cline{3-5} 
 &  & 3 & 73.39\% & 73.11\% \\ \cline{3-5} 
 &  & 4 & 72.38\% & 71.15\% \\ \cline{3-5} 
 &  & 5 & 72.31\% & 72.00\% \\ \cline{3-5} 
 &  & \textbf{Average} & \textbf{72.16\%} & \textbf{71.95\%} \\ \cline{2-5} 
 & \multirow{6}{*}{\textbf{300}} & 1 & 75.77\% & 73.93\% \\ \cline{3-5} 
 &  & 2 & 75.00\% & 71.47\% \\ \cline{3-5} 
 &  & 3 & 70.23\% & 69.73\% \\ \cline{3-5} 
 &  & 4 & 75.67\% & 75.95\% \\ \cline{3-5} 
 &  & 5 & 71.35\% & 70.12\% \\ \cline{3-5} 
 &  & \textbf{Average} & \textbf{73.60\%} & \textbf{72.24\%} \\ \cline{2-5} 
 & \multirow{6}{*}{\textbf{500}} & 1 & 73.90\% & 74.24\% \\ \cline{3-5} 
 &  & 2 & 76.33\% & 76.49\% \\ \cline{3-5} 
 &  & 3 & 73.96\% & 73.48\% \\ \cline{3-5} 
 &  & 4 & 73.03\% & 74.30\% \\ \cline{3-5} 
 &  & 5 & 71.35\% & 72.86\% \\ \cline{3-5} 
 &  & \textbf{Average} & \textbf{73.71\%} & \textbf{74.27\%} \\ \hline
\multirow{18}{*}{\textbf{Fashion-MNIST}} & \multirow{6}{*}{\textbf{100}} & 1 & 79.96\% & 79.96\% \\ \cline{3-5} 
 &  & 2 & 78.77\% & 77.63\% \\ \cline{3-5} 
 &  & 3 & 78.76\% & 78.82\% \\ \cline{3-5} 
 &  & 4 & 80.48\% & 79.60\% \\ \cline{3-5} 
 &  & 5 & 78.77\% & 79.92\% \\ \cline{3-5} 
 &  & \textbf{Average} & \textbf{79.35\%} & \textbf{79.19\%} \\ \cline{2-5} 
 & \multirow{6}{*}{\textbf{300}} & 1 & 79.93\% & 80.07\% \\ \cline{3-5} 
 &  & 2 & 80.56\% & 80.92\% \\ \cline{3-5} 
 &  & 3 & 80.49\% & 81.10\% \\ \cline{3-5} 
 &  & 4 & 79.72\% & 80.69\% \\ \cline{3-5} 
 &  & 5 & 78.04\% & 79.24\% \\ \cline{3-5} 
 &  & \textbf{Average} & \textbf{79.75\%} & \textbf{80.40\%} \\ \cline{2-5} 
 & \multirow{6}{*}{\textbf{500}} & 1 & 80.67\% & 81.64\% \\ \cline{3-5} 
 &  & 2 & 79.55\% & 80.06\% \\ \cline{3-5} 
 &  & 3 & 77.83\% & 81.73\% \\ \cline{3-5} 
 &  & 4 & 79.47\% & 79.13\% \\ \cline{3-5} 
 &  & 5 & 80.05\% & 80.85\% \\ \cline{3-5} 
 &  & \textbf{Average} & \textbf{79.51\%} & \textbf{80.68\%} \\ \hline
\end{tabular}}
\caption{Results of MNIST, KMNIST and Fashion-MNIST with 400 seeds.}
\label{400seeds}
\end{table}

\begin{table}[!htb]
\centering
\scalebox{0.8}{
\begin{tabular}{|c|c|c|c|c|}
\hline
\textbf{Dataset} & \textbf{Epochs} & \textbf{Experiment} & \textbf{Accuracy} & \textbf{Accuracy (Best Model)} \\ \hline
\multirow{18}{*}{\textbf{MNIST}} & \multirow{6}{*}{\textbf{100}} & 1 & 96.73\% & 96.33\% \\ \cline{3-5} 
 &  & 2 & 96.79\% & 96.60\% \\ \cline{3-5} 
 &  & 3 & 96.73\% & 96.10\% \\ \cline{3-5} 
 &  & 4 & 95.97\% & 96.09\% \\ \cline{3-5} 
 &  & 5 & 95.11\% & 95.40\% \\ \cline{3-5} 
 &  & \textbf{Average} & \textbf{96.27\%} & \textbf{96.10\%} \\ \cline{2-5} 
 & \multirow{6}{*}{\textbf{300}} & 1 & 98.01\% & 97.83\% \\ \cline{3-5} 
 &  & 2 & 97.25\% & 97.34\% \\ \cline{3-5} 
 &  & 3 & 96.80\% & 96.99\% \\ \cline{3-5} 
 &  & 4 & 97.05\% & 97.33\% \\ \cline{3-5} 
 &  & 5 & 97.10\% & 97.35\% \\ \cline{3-5} 
 &  & \textbf{Average} & \textbf{97.24\%} & \textbf{97.37\%} \\ \cline{2-5} 
 & \multirow{6}{*}{\textbf{500}} & 1 & 98.00\% & 98.03\% \\ \cline{3-5} 
 &  & 2 & 96.68\% & 96.66\% \\ \cline{3-5} 
 &  & 3 & 96.68\% & 96.66\% \\ \cline{3-5} 
 &  & 4 & 97.64\% & 97.64\% \\ \cline{3-5} 
 &  & 5 & 97.54\% & 97.74\% \\ \cline{3-5} 
 &  & \textbf{Average} & \textbf{97.31\%} & \textbf{97.35\%} \\ \hline
\multirow{18}{*}{\textbf{KMNIST}} & \multirow{6}{*}{\textbf{100}} & 1 & 73.53\% & 72.52\% \\ \cline{3-5} 
 &  & 2 & 72.78\% & 69.95\% \\ \cline{3-5} 
 &  & 3 & 75.81\% & 75.73\% \\ \cline{3-5} 
 &  & 4 & 70.35\% & 71.73\% \\ \cline{3-5} 
 &  & 5 & 75.72\% & 73.38\% \\ \cline{3-5} 
 &  & \textbf{Average} & \textbf{73.64\%} & \textbf{72.66\%} \\ \cline{2-5} 
 & \multirow{6}{*}{\textbf{300}} & 1 & 73.52\% & 73.57\% \\ \cline{3-5} 
 &  & 2 & 73.71\% & 74.00\% \\ \cline{3-5} 
 &  & 3 & 75.72\% & 76.61\% \\ \cline{3-5} 
 &  & 4 & 74.02\% & 74.97\% \\ \cline{3-5} 
 &  & 5 & 77.03\% & 77.59\% \\ \cline{3-5} 
 &  & \textbf{Average} & \textbf{74.80\%} & \textbf{75.35\%} \\ \cline{2-5} 
 & \multirow{6}{*}{\textbf{500}} & 1 & 74.88\% & 75.43\% \\ \cline{3-5} 
 &  & 2 & 75.27\% & 74.51\% \\ \cline{3-5} 
 &  & 3 & 73.70\% & 72.99\% \\ \cline{3-5} 
 &  & 4 & 76.76\% & 78.11\% \\ \cline{3-5} 
 &  & 5 & 75.25\% & 75.77\% \\ \cline{3-5} 
 &  & \textbf{Average} & \textbf{75.17\%} & \textbf{75.36\%} \\ \hline
\multirow{18}{*}{\textbf{Fashion-MNIST}} & \multirow{6}{*}{\textbf{100}} & 1 & 80.55\% & 80.21\% \\ \cline{3-5} 
 &  & 2 & 81.58\% & 82.40\% \\ \cline{3-5} 
 &  & 3 & 78.93\% & 79.84\% \\ \cline{3-5} 
 &  & 4 & 82.08\% & 81.66\% \\ \cline{3-5} 
 &  & 5 & 80.53\% & 80.40\% \\ \cline{3-5} 
 &  & \textbf{Average} & \textbf{80.73\%} & \textbf{80.90\%} \\ \cline{2-5} 
 & \multirow{6}{*}{\textbf{300}} & 1 & 79.87\% & 81.03\% \\ \cline{3-5} 
 &  & 2 & 80.09\% & 81.41\% \\ \cline{3-5} 
 &  & 3 & 81.81\% & 82.45\% \\ \cline{3-5} 
 &  & 4 & 81.75\% & 82.82\% \\ \cline{3-5} 
 &  & 5 & 80.19\% & 79.92\% \\ \cline{3-5} 
 &  & \textbf{Average} & \textbf{80.74\%} & \textbf{81.53\%} \\ \cline{2-5} 
 & \multirow{6}{*}{\textbf{500}} & 1 & 80.69\% & 81.46\% \\ \cline{3-5} 
 &  & 2 & 79.65\% & 79.84\% \\ \cline{3-5} 
 &  & 3 & 80.20\% & 81.13\% \\ \cline{3-5} 
 &  & 4 & 80.27\% & 80.28\% \\ \cline{3-5} 
 &  & 5 & 79.79\% & 79.28\% \\ \cline{3-5} 
 &  & \textbf{Average} & \textbf{80.12\%} & \textbf{80.40\%} \\ \hline
\end{tabular}}
\caption{Results of MNIST, KMNIST and Fashion-MNIST with 500 seeds.}
\label{500seeds}
\end{table}

\subsection{Seed Images used in experiments of RQ3}\label{rq3appendix}
\begin{figure}[htb!]
    \centering
    \scalebox{1.1}{
    \includegraphics{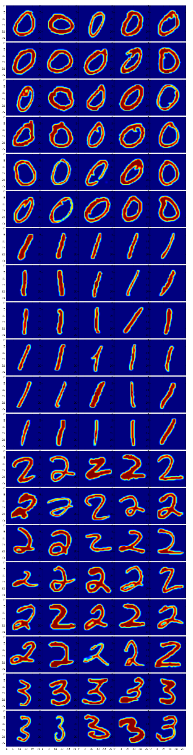}}
    \caption{Seed \textit{seed\_14} used in MNIST experiment as part of RQ3}
    \label{mnistrq3_img}
\end{figure}

\begin{figure}[!htb]
\centering
\scalebox{1.1}{
    
    \includegraphics{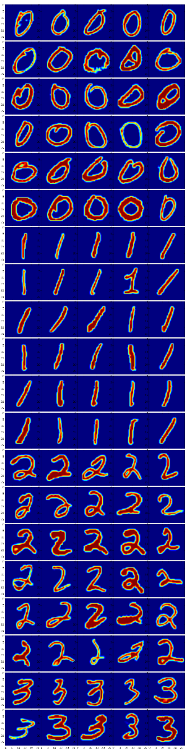}}
    \caption{Seed \textit{seed\_102} used in MNIST experiment as part of RQ3}
    \label{mnistrq3_img2}
\end{figure}

Figures \ref{mnistrq3_img}-\ref{mnistrq3_img2}, shows seeds that produce highest and lowest results for MNIST datasets in RQ3.

\end{document}